\documentclass{article}

\usepackage{PRIMEarxiv}

\usepackage[utf8]{inputenc} 
\usepackage[T1]{fontenc}    
\usepackage{hyperref}       
\usepackage{url}            
\usepackage{booktabs}       
\usepackage{amsfonts}       
\usepackage{nicefrac}       
\usepackage{microtype}      
\usepackage{lipsum}
\usepackage{graphicx}       
\usepackage{multirow}
\usepackage{tabularx}%
\usepackage{lscape}
\usepackage{amsmath}
\usepackage{amssymb}
\usepackage{subcaption}
\usepackage{makecell}
\usepackage{titlesec}
\usepackage{float}

\usepackage{fancyhdr}       

\newcolumntype{Y}{>{\centering\arraybackslash}X}

\usepackage{algorithm}%
\usepackage{algorithmicx}%
\usepackage[noend]{algpseudocode}%

\DeclareMathOperator*{\argmax}{argmax}

\newcommand{\LineComment}[1]{\hfill\textit{#1}}

\title{Parallel Hyperparameter Optimization Of Spiking Neural Networks
\thanks{Submitted to Neurocomputing} 
}

\author{
  Thomas Firmin \\
    CNRS, Inria, Centrale Lille, UMR 9189 CRIStAL\\
    Université de Lille\\
    Lille, F-59000 France \\
  \texttt{thomas.firmin@univ-lille.fr} \\
   \And
   Pierre Boulet \\
    CNRS, Inria, Centrale Lille, UMR 9189 CRIStAL\\
    Université de Lille\\
    Lille, F-59000 France \\
  \texttt{pierre.boulet@univ-lille.fr} \\
  \And
  El-Ghazali Talbi \\
    CNRS, Inria, Centrale Lille, UMR 9189 CRIStAL\\ 
    Université de Lille\\
    Lille, F-59000 France \\
  \texttt{el-ghazali.talbi@univ-lille.fr} \\
}

\begin{document}
\maketitle

\begin{abstract}
Hyperparameter optimization of spiking neural networks (SNNs) is a difficult task which has not yet been deeply investigated in the literature. In this work, we designed a scalable constrained Bayesian based optimization algorithm that prevents sampling in non-spiking areas of an efficient high dimensional search space. These search spaces contain infeasible solutions that output no or only a few spikes during the training or testing phases, we call such a mode a  “silent network”. Finding them is difficult, as many hyperparameters are highly correlated to the architecture and to the dataset. We leverage silent networks by designing a spike-based early stopping criterion to accelerate the optimization process of SNNs trained by Spike Timing Dependent Plasticity (STDP) and surrogate gradient. We parallelized the optimization algorithm asynchronously, and ran large-scale experiments on heterogeneous multi-GPU Petascale architecture. Results show that by considering silent networks, we can design more flexible high-dimensional search spaces while maintaining a good efficacy. The optimization algorithm was able to focus on networks with high performances by preventing costly and worthless computation of silent networks.
\end{abstract}



\section{Introduction}

Spiking Neural Networks (SNNs) are analog networks embracing a more biologically inspired and event-based approach compared to their counterpart, traditional Artificial Neural Networks (ANNs). The distinctive characteristics of SNNs, such as in-memory and local computations, make them impractical for execution on usual von Neumann architectures. Instead, they are computed on specialized neuromorphic hardware platforms, such as SpiNNaker~\cite{spinnaker} or Loihi~\cite{loihi}. SNNs use time as a resource for computations and communications~\cite{maas}. These specificities make SNNs energy-efficient, massively scalable and parallelizable due to the theoretical asynchronicity between the neurons. SNNs rely on the dynamics between spikes within the network. The choice of the neuron model, the architecture, and the learning rule, all affect these dynamics. Therefore, SNNs have their own hyperparameters (HPs), use cases and datasets~\cite{neurobench, schuman}. One of today's challenges, is the  training algorithms, used to find optimal synaptic weights. Three main approaches are available, ANN-to-SNN conversion, plasticity rules or surrogate gradient backpropagation. Here, we decided to address HyperParameter Optimization (HPO) of SNNs, by using two training algorithms, STDP~\cite{surveystdp} plasticity rule and a state-of-the-art surrogate gradient-based method known as SLAYER~\cite{slayer}.

HyperParameter Optimization (HPO) is a well-studied problem in machine learning~\cite{hpo_bischl, talbi} and tuning of SNNs' HPs also falls into this category of problems.
We consider a high dimensional mixed and constrained mono-objective HPO problem, to achieve the best possible accuracy on classification tasks. Hence, we define the HPO of a SNN $\mathcal{N}$ by:
\begin{equation}
    \lambda^* \in \argmax_{\lambda \in \Lambda} \mathcal{L}(\mathcal{N}_{\theta^*}^{\lambda}, \mathcal{D}_{\text{valid}})\text{, s.t. } c_1 \leq 0, c_2 \leq 0,...
\end{equation}
where $\Lambda$ is the search space containing all HPs combinations $\lambda$, and $\lambda^*$ is the theoretical optimum. The loss function $\mathcal{L}$ denotes the classification error, to be maximized by the optimization algorithm. The loss $\mathcal{L}$ is computed on the classification of a validation dataset $\mathcal{D}_{valid}$ made by a trained SNN $\mathcal{N}$ under $\lambda$, a fixed combination of HPs.  The black-box constraints must be of the form $c_i \leq 0$.
The optimal parameters $\theta^*$ of a network $\mathcal{N}$ are determined on a training dataset $\mathcal{D}_{\text{train}}$. The training process is written by: $\theta^* \in \argmax_{\theta \in \Theta}\mathcal{L}(\mathcal{N}_{\theta}^{\lambda},D_{\text{train}})$. The final accuracy is assessed on an untouched testing dataset $\mathcal{D}_{\text{test}}$: $Accuracy(\mathcal{N^{\lambda^*}_{\theta^*}}, \mathcal{D}_{\text{test}})$.
\newline\newline
The problem is characterized by a black-box objective function, i.e. the training and validation of a SNN. The most challenging property is the expensiveness and randomness in terms of the computation time of a single solution, which can last from minutes to tens of hours. Common optimization algorithms, e.g. Grid-Search (GS), Evolutionary Algorithm (EA) or Bayesian Optimization (BO) used for HPO of a wide range of Machine Learning models, have been extended to SNNs. Most of the proposed search spaces are restrained to a few HPs with heavily restricted bounds~\cite{parsa_bo, parsa_multi}. Indeed, designing high dimensional search spaces integrating many HPs introduces challenging problems. The main issue is non-spiking areas. These areas of the search space contain infeasible solutions, as SNNs output an insufficient number of spikes. We called these solutions “silent networks”.
\newline\newline
The main contributions of the paper are as follows:
\begin{itemize}
\item The design of efficient high dimensional search spaces containing silent networks. We leverage silent networks by designing a spike-based early stopping criterion and black-box constraints to prevent unnecessary computations.
\item A scalable Bayesian based optimization algorithm for high dimensional constrained search spaces to avoid sampling non-spiking areas. We have also generalized this approach to the two most popular families of training algorithms, plasticity rules and surrogate gradient.
\item A parallel asynchronous implementation has been developed. Design of experiments was carried out on clusters of GPUs (Grid'5000~\cite{grid5000}). Large-scale experiments were conducted on heterogeneous multi-GPU Petascale architecture (Jean Zay supercomputer).
\end{itemize}

The paper is organized as follows. In Section~\ref{sec:works} we review some HPO of SNNs within the existing literature. Subsequently, in Section~\ref{sec:early_stopping}, we present the main contributions of this paper. We leverage “silent networks” to design a spike-based early stopping criterion and black-box constraints to accelerate the exploration of high dimensional search spaces. In Section~\ref{sec:exp}, different experimental setups are presented in terms of SNN architecture, optimization algorithm, search space design and datasets. Computational results about the overall HPO process and accuracy are then discussed in Section~\ref{sec:results}. Finally, we present the main conclusions of this work and some perspectives related to HPO of SNNs in Section~\ref{sec:con}.

\section{Related works}\label{sec:works}

Designing the search space and selecting decision variables are crucial steps in HPO~\cite{hpo1, hpo2, talbi}. Hyperparameters can be of various types; continuous, discrete and categorical. The major difficulty is to select the most relevant ones and to define their bounds. This selection is crucial, as it is the step, during which, one has to inject knowledge in the optimization process.

In SNNs, the hyperparameters values can be defined according to biology or manually~\cite{georges, yu, izhikevitch, hh}. These initial combinations of hyperparameters, are often used as a baseline to compare the efficacy of the HPO process~\cite{hpo1}. However, these “off-the-shelf” combinations are not universal and not suitable for all problems and datasets~\cite{nflt_snn}. Performing HPO can result in unthinkable solutions compared to usually handcrafted ones~\cite{nasnn1}. Hyperparameters of SNNs are known to be very sensitive~\cite{chakraborty, guo}, so we have to integrate this specific knowledge to improve high dimensional HPO. Moreover, parallel HPO of SNNs on heterogeneous multi-nodes and multi-GPUs architectures is to our knowledge unknown.

\subsection{Spike Timing Dependent Plasticity based SNNs}

In~\cite{chernyshev} 7 hyperparameters were optimized using a sequential Surrogate-based optimization and applied to a Spiking Reservoir made of Spike Response Model (SRM). The training is carried out by STDP. Thus, HPs of the training algorithm are optimized, such as both traces time constant ($\tau_-$, $\tau_+$) and ratios of long-term potentiation and depression ($A_+$, $A_-$). One HP is linked to the neuron model and three others to the network topology (e.g. initial inhibitory and excitatory strength). The problem is a classification task and the dataset is made of artificially generated time series. Experiments indicate that the initial value of inhibitory weights, and the hyperparameter of the neuron model have a greater impact on the accuracy than learning rates. However, no analysis of the spiking activity is made to explain these results. The authors suggest a more advanced study and the use of a better BO technique.

Evolutionary algorithms are also applied to such problems. In~\cite{polap}, the authors optimized the Diehl \& Cook self-organizing map architecture~\cite{diehl} trained by STDP in a federated learning environment. The optimized search space is made of 6 HPs: number of neurons, excitatory strength, inhibitory strength, neurons' threshold, threshold potential decay and $\mu$ a STDP coefficient. The authors applied 5 different metaheuristics, Cuckoo Search, Whale Optimization, Polar Bears, Grasshopper Optimization and Salp Swarm algorithm. The best accuracies found were respectively $89\%$ and $82\%$ on MNIST and Fashion MNIST.
Similarly, Guo et al.~\cite{guo},  optimized 5 HPs, $\theta^+$ for the threshold adaptation, both STDP learning rates and both pre- and post-synaptic traces time constants, achieving 86.54\% accuracy on Poisson encoded MNIST using the Dielh \& Cook architecture, a Genetic Algorithm (GA) and a fitness function considering accuracy and latency. They show that time constants with high values give higher accuracies but slower convergence, and conversely for small values. The authors also studied the impact of the numerical methods (Euler and third order Runge-Kutta), on the accuracy and FPGA implementation. They also pointed out the high sensitivity of SNNs to their HPs, and the computational complexity of simulated SNNs, with training taking up to 11 hours, for a simple two layers SOM architecture.

In~\cite{shahsavari}, the authors did not perform an HPO, but they manually tuned, 1 HP, the number of neurons. Experimental results have demonstrated the relationship between accuracy and the number of output neurons of a Restricted Boltzmann Machine (RBM) trained with STDP. The higher the number of neurons, the better the accuracy. They achieved a maximum of $89,4\%$ of accuracy on MNIST.

HPO was applied to Locust Lobula Giant Movement Detector (LGMD) model trained by STDP, in~\cite{salt}. The authors were able to optimize up to 18 different hyperparameters of 4 different versions of a given architecture implemented using Brian2\cite{brian2}. They applied these models to the detection of looming and nonlooming stimuli on an in-situ DVS recorded UAV dataset. The applied optimization algorithms were random search, three BO techniques, Differential Evolution (DE) and Self-adaptive DE.

Classifying DVS-Gesture using STDP is a tough task, notably due to the spatio-temporal features and to the heterogeneous distribution of spikes among classes (See Figure \ref{fig:dvs_dist}). In~\cite{georges}, a Convolutional Spiking Neural Network (CSNN) combined with a reservoir was used. Outputs are decoded using a softmax  regression. The authors compared different networks with hand-tuned hyperparameters linked to the architecture and topology. STDP, neuron model or homeostasis hyperparameters are fixed. They achieved a $65\%$ of accuracy with a network of $3.1764$ millions of parameters.
DVS-Gesture classification was also tackled in~\cite{iyer}, by using a combination of unsupervised STDP and supervised Tempotron on a SOM like architecture. The model achieved $60.37\%$ of accuracy by using an augmented dataset. The authors manually studied the impact of the inhibition strength. The number of neurons was also studied, and it appears that training with augmented data and large number of neurons gives the best accuracies.
Achieving such accuracies on DVS-Gesture using exclusively unsupervised STDP and non Machine Learning-based decoders (max or average spikes) is to our knowledge unknown.

\subsection{Gradient based SNNs}

Bayesian Optimization HyperBand (BOHB) was used in~\cite{vicente} to find the optimal solution between 3 HPs: leakage, time-steps and learning rate. The methodology was applied on CIFAR-100 using S-ResNET38. BOHB~\cite{bohb} is a multi-fidelity state-of-the-art algorithm for HPO~\cite{hpo1}. It combines Bayesian optimization~\cite{garnett} and Successive halving~\cite{hyperband}. The authors also did a comparison between accuracy and sample duration. They have tested 10, 20, 30, 40 and 50 frames, and showed that the leakage factor of the LIF and number of frames are linked. Empirical results show that by reducing the number of frames, one can adapt the leakage to counterbalance the accuracy loss due to shorter samples. Moreover, other works based on surrogate gradient show the impact of the neurons' threshold on the accuracy~\cite{thresh_imp}.

In \cite{parsa_slayer}, the authors optimized an 8 HPs discretized search space of size 512, by using Hierarchical-Bayesian HPO Algorithm (H-PABO). Four convolutional architectures can be selected during the optimization, and these are trained with SLAYER, on the DVS Gesture dataset. Accuracy and latency were optimized and the best solutions according to the two objectives are 95.113\%, 73.17sec and 91.35\%, 29.19sec.

HPO was also applied on a 3 layers CSNN trained by NormAD~\cite{kulkarni}. They obtained a 98.17\% of accuracy on MNIST, where pixel intensities are directly converted into input current. Authors manually optimized images' presentation time, number of convolution filters and learning rates. For current encoded MNIST they empirically show that a 100ms exposition time is optimal. Lower and higher values result in lower accuracies.

Energy consumption and spiking activity are also a matter for gradient based SNNs. In~\cite{cordone}, a Grid Search was carried out to fix the parameter $\alpha$ used to approximate the gradient of the Heaviside function. The authors studied the impact of sparse convolutions on the internal spiking activity of the network on the DVS128 Gesture dataset. They have also investigated the impact of the sample duration on the accuracy, and concluded that the longer the sample duration, the better the accuracy, until a certain value after which accuracy remains almost constant. They also mentioned an interesting behavior of the neuron's threshold HP on zero-spiking layers. For some values of the threshold, a layer may become completely silent.

\subsection{Other approaches}

Evolutionary Optimization for Neuromorphic Systems (EONS)~\cite{eons1} uses evolutionary optimization to tune parameters and hyperparameters of SNNs. Here, the application is strongly linked to the neuromorphic hardware (Caspian neuromorphic computing system). In~\cite{parsa_bo}, up to 11 HPs were optimized within a discretize search space of maximum size 54 432 000. Experiments were carried out on EONS by using BO and Grid Search, authors show that BO performs better than GS. They have empirically demonstrated that HPs of the encoding method and hardware, have a greater impact than HPs linked to the GA used to train SNNs. The same authors~\cite{parsa_multi}, applied H-PABO, a multi-objective BO, on 11 discrete HPs to optimize an EONS trained architecture on two and three objectives: network performances, energy consumption and number of synapses.

\section{Infeasible solutions, early stopping and black-box constraints}\label{sec:early_stopping}

\subsection{Silent networks and infeasible solutions}

One of the major challenges of SNNs is the signal loss problem, induced by a decreasing firing rate through layers~\cite{info_loss, kim_deep}. This phenomenon limits the depth of SNNs since too deep SNNs might result in networks unable to output spikes~\cite{cordone}. In this work, we extend this problem to deep and shallow networks outputting an insufficient spiking activity to perform a specific task. We call these infeasible solutions “silent networks”, as the lack of spiking activity is not only explained by the depth of the network but also by mistuned HPs or architecture.

Explaining silent networks by only considering the depth of a network is misleading, especially when performing HPO. Indeed, one can imagine a binary classification problem in which, no output spikes corresponds to class 0, output spikes to class 1. The input data is made of spikes, and contains both classes. One can solve this problem using a shallow network of two layers (inputs and outputs) made of LIF neurons, with a resting potential set to 0. Then, the neurons' threshold can be set to infinity because nothing prohibits this. Considering a finite number of input spikes, neurons, and synapses with finite weights, then we obtain a shallow network unable to spike – a silent network. This network is an \textit{infeasible solution}, as output spikes were expected.

We just framed one challenge of HPO, which is designing a viable search space, and finding bounds of the decision variables. In the previous thought experiment, the lower bound of the threshold is straightforward, while the upper one is more challenging. To define it, information about the inputs, topology, spike frequency or other HPs is needed. Even with this information, a question remains: does the threshold itself solely explain the silent network?
We can extend this question to broader architectures, neuron's models and HPs.

\begin{figure*}[ht]
    \centering
    \includegraphics[width=\textwidth]{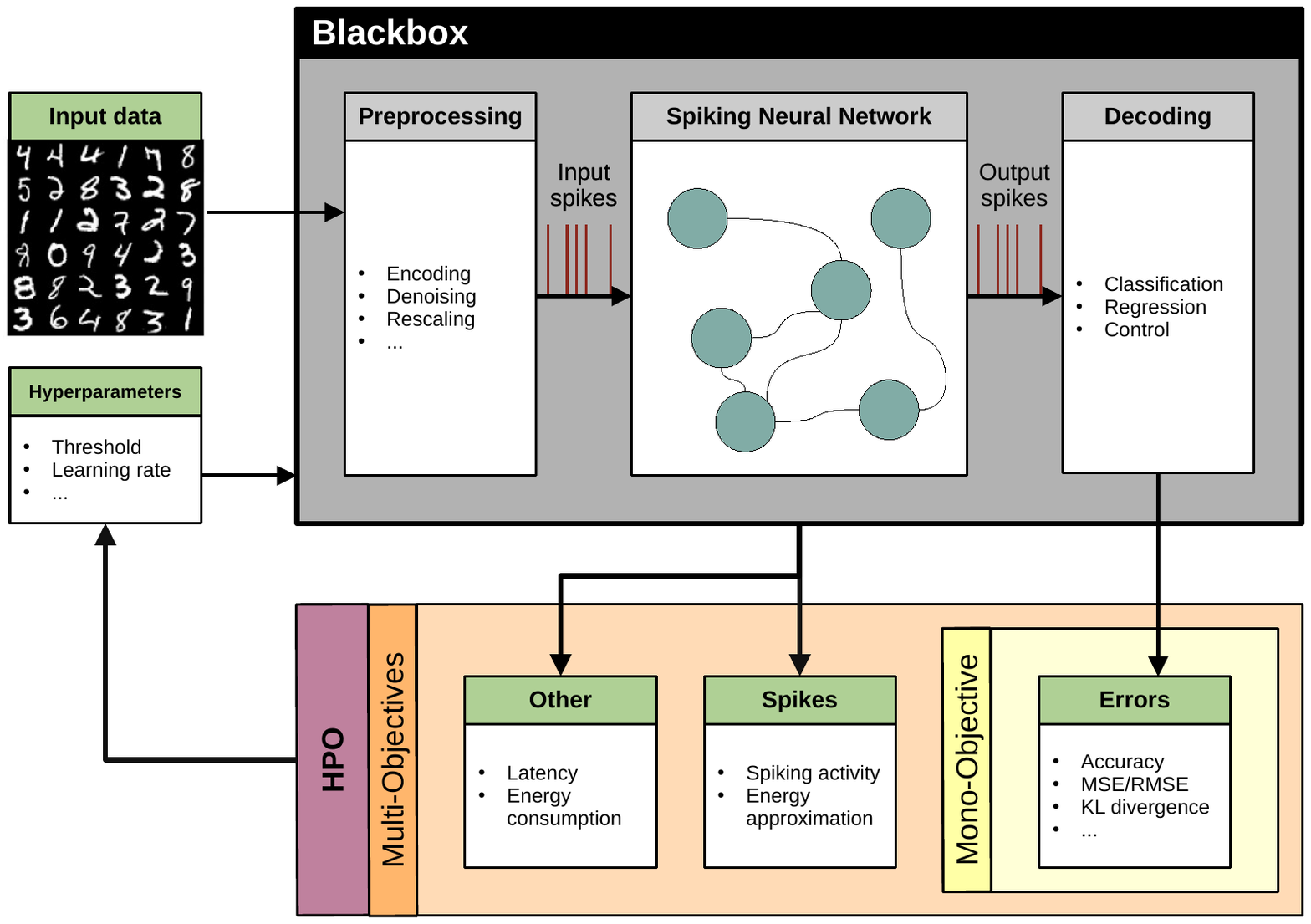}
    \caption{Workflow of a usual mono- or multi-objectives HPO of SNNs.}
    \label{fig:hpo_process}
\end{figure*}

In a mono- or multi-objectives HPO, a SNN, its training, validation, computations of the errors and other information, are considered as fully black-box. The optimization algorithm has only access to the outputs of this blackbox. This process is described in figure \ref{fig:hpo_process}. In a classification task, the output of the blackbox is usually the accuracy of the network on the validation dataset.
In this case, the decoder (e.g. average spike or max spike) becomes a trap for HPO. Indeed, the decoder can create a link between non-spiking outputs and a class, value or action. If we look at the previous binary classification problem with a silent network, then the average accuracy will be about $50\%$ (random).
The error, here, is to consider a $50\%$ accuracy silent network, similar to a $50\%$ accuracy spiking network. This is what the HPO algorithm is doing. Because of the blackbox, spiking and non-spiking networks with equal objectives are considered the same.

Moreover, in many works, the minimization of the energy consumption of a network is tackled by considering the spiking activity within a multi-objectives context where the number of spikes or number of neurons has to be minimized~\cite{parsa_multi, kim, dimovska}. The trap in this context is more vicious. Here, a silent network, doing nothing, can be considered better than a low spiking one.

A simple solution, to avoid silent networks during HPO, would be to sufficiently restrain the bounds of the search space to only consider non-silent networks. However, by doing so, one forgets the complex correlations of HPs with accuracy and spiking activity. Indeed, by strongly restraining the search space, one can reject many suitable solutions. This becomes even harder in high dimensional HPO for deep SNNs. This problem is even more crucial and difficult when energy consumption or spiking activity has to be minimized. Indeed, one needs to find the frontier between silent and low-spiking networks.

\subsection{Early stopping}

A way to accelerate HPO process is to early detect silent networks and avoid useless expensive computations. Indeed, in the current workflow, no matter if a network is spiking or not, samples are presented until the whole dataset is processed, and until all epochs are computed.  No matter the outputs, the equations of neurons and learning algorithm are still computed. This phenomenon gets worse when the network is trained with Hebbian based rules, if cells do not fire, then synapses do not wire together.

\begin{figure}[ht]
     \centering
     \begin{subfigure}[t]{0.49\textwidth}
         \centering
    \includegraphics[trim={0 0 0 0.7cm},clip,width=\linewidth]{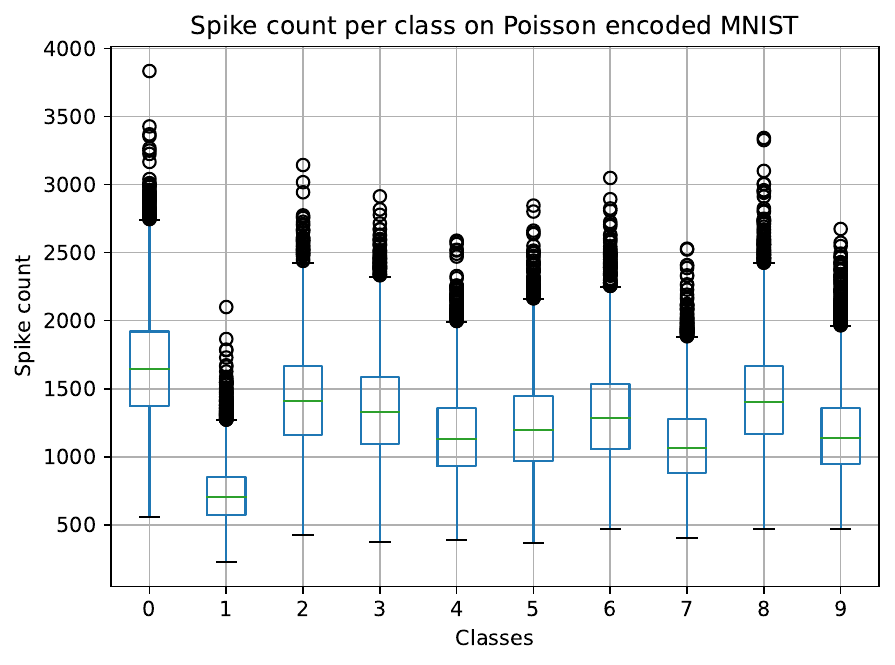}
         \caption{100 frames Poisson encoded MNIST}
         \label{fig:mnist_dist}
     \end{subfigure}
     \begin{subfigure}[t]{0.49\textwidth}
         \centering
         \includegraphics[trim={0 0 0 0.7cm},clip,width=\textwidth]{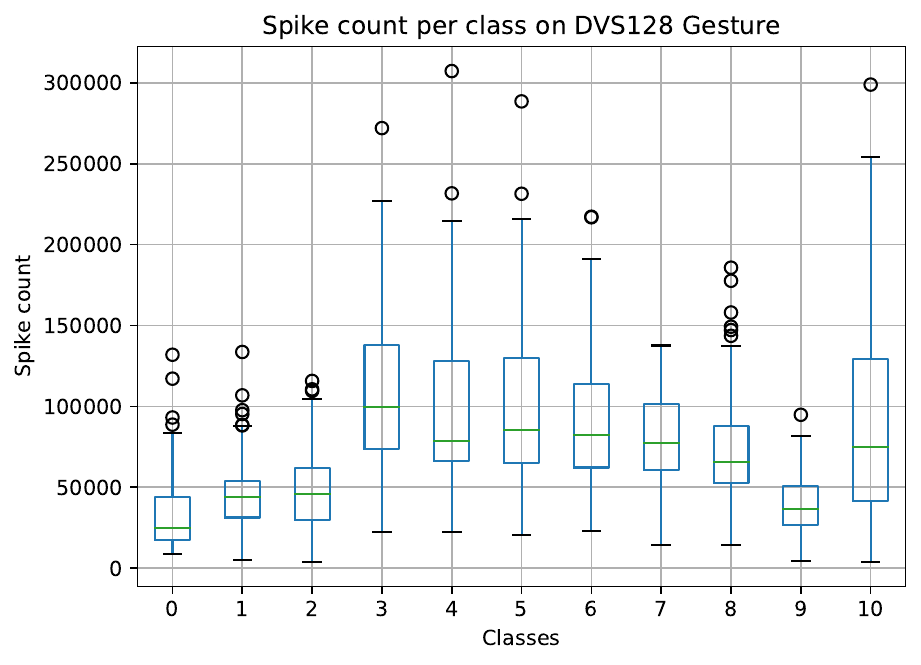}
         \caption{100 frames preprocessed DVS128 Gesture}
         \label{fig:dvs_dist}
     \end{subfigure}
    \caption{Class wise spikes count distribution}
    \label{fig:class_dist}
\end{figure}

In this work, we define an early stopping criterion based on the percentage of images that did not emit at least $\alpha$ spikes, during the presentation of an image. A network can sometimes be only active for a few data since it can be made of imbalanced number of spikes. In figure \ref{fig:mnist_dist}, one can see that class 1 of MNIST produces fewer spikes than others. Figure \ref{fig:dvs_dist} shows that for DVS128 Gesture, these disparities are more pronounced and are seen among data of a same class. For example, wider or narrower gestures produce more or less spikes. The training pipeline of a SNNs is described in algorithm \ref{alg:stop}, it can be easily extended to epochs and batch of larger size.
This early stopping, allows interrupting the training if a certain percentage $\beta$ of data does not emit at least $\alpha$ spikes. For example, during the training phase, if at least $\beta = 5\%$ of the data have not emitted at least $\alpha=1$ spike when presented to the network, training is interrupted.

We can extend this stopping criterion to other layers, such as the inhibitory layer in a Diehl \& Cook architecture~\cite{diehl}. This layer enables neurons from the excitatory layer to specialize into a certain spiking pattern. Thus, to learn, the architecture requires spikes at both the output and at the inhibitory layers.

The $\alpha$ and $\beta$ HPs can be set according to input data, architecture, or training algorithm. For instance, in SLAYER, the outputs require a certain rate of spikes and the backpropagation can enforce the network to spike. Whereas in a STDP trained SOM, if there are no output spikes, then there is no training nor weights update.
So for surrogate gradient, $\beta$ can be greater than for networks trained by STDP. For SNNs trained by SLAYER, $\alpha$ can be set according to the number of expected output spikes, with a certain tolerance, since the gradient can enforce spiking activity. It is also important to define $\beta$ according to the training time. Spending time training a silent network is time that cannot be spent on training promising networks with minimum spiking activity.

We now have of more general definition of what silent networks are, depending on the values of $\alpha$ and $\beta$. The objective is to prevent a network of being stopped during training because of a lack of spiking activity on certain data.

{
\begin{figure}
    \centering
    \begin{minipage}{\linewidth}
        \begin{algorithm}[H]
            \caption{SNN training pipeline with early stopping for 1 epoch and no batch}\label{alg:stop}
            \begin{algorithmic}[1]
                \Require{\\
                    $\mathcal{N}$ \LineComment{Network}\\
                    $X_{\text{train}}$ \LineComment{Training data} \\
                    $Y_{\text{train}}$ \LineComment{Training labels}\\
                    $S$ \LineComment{Number of samples}\\
                    $\alpha$ \LineComment{Minimum spiking activity}\\
                    $\beta$ \LineComment{Maximum of non spiking data}\\
                }
                \Ensure{$\mathcal{N}$ \LineComment{Trained network}}
                \State $i \gets 1$
                \State $count \gets 0$ \LineComment{Number of non spiking data}
                \State $out \gets \emptyset$ \LineComment{Output spikes}
                \While{$(count/S \leq \beta) \land (i \leq S)$}
                \State $out \gets \text{Train}(\mathcal{N}, X_{\text{train}}[i], Y_{\text{train}}[i])$
                \If{$\texttt{SUM}(out) < \alpha$}
                \State $count \gets count + 1$
                \EndIf
                \State $i \gets i + 1$
                \EndWhile
                \Return $\mathcal{N}$
            \end{algorithmic}
        \end{algorithm}
    \end{minipage}
\end{figure}
}

\subsection{Blackbox constraints}

The early stopping criterion allows detecting silent networks, and so prevents useless and costly computations. By spending less time on silent networks, the HPO can evaluate more networks within a given runtime. However, we can improve this by considering silent networks within the HPO process. Indeed, to accelerate the exploration of the search space, one could forecast the spiking activity of a network to avoid sampling within areas of the search space containing silent networks. However, predicting the exact spiking activity of a SNNs is in practice difficult. We can only get the exact spiking activity by passing data to the network and retrieving its outputs.

There are various means to handles constraints in HPO, rejecting, penalizing, repairing or preserving~\cite{talbi}.
Because of the stochasticity and unpredictability of the spiking activity of SNNs, repairing and preserving strategies are ruled out. Therefore, constraints on the spiking activity are blackbox as we cannot formally model them.
In our specific context, it is not advisable to systematically reject silent networks. Indeed, a network that has been early stopped does not necessarily mean that it as low performances. Multi-fidelity shows~\cite{hazan} that we can obtain good performances with a reduced dataset. Thus, the accuracy of a silent network that does not follow constraints is still computed on the validation dataset, as some samples of the data might output spikes. 

Thus, we apply a penalization by rewriting the early stopping to create a violation value:
\begin{equation}\label{eq:constraints}
 \sum\limits_{c \in \mathcal{C}}max\left(\frac{count_c}{S}-\beta_c, 0\right)   
\end{equation}
Where, $S$ is the total number of samples, $\mathcal{C}$ is a set of constraints on different layers. If $\exists c \in \mathcal{C} : \frac{count_c}{S} > \beta_c$ for at least one epoch, the training is interrupted and penalized by at least a positive value, which slightly varies depending on the number of non-spiking data within batches and update interval of the stopping criterion. So, when constraints are met, it describes an acceptable proportion of silent data during training. Thanks to black-box constraints, one of the objectives of the HPO is to avoid sampling silent networks, so to prevent the training of being stopped because of a lack of spiking activity.

\section{Experimental setup}\label{sec:exp}

\subsection{Optimization algorithm}

The goal of this work is to compare HPO of SNNs trained by STDP and SLAYER, i.e., with different degrees of biological-plausibility. In section \ref{sec:early_stopping}, we described what silent networks are. We designed a spike-based early stopping criterion, and associated black-box constraints, to avoid sampling these silent networks. Training SNNs takes time, from minutes to hours. Some preliminary experiments, using a simple GA and Steady-state GA, failed due to the high number of evaluations necessary to make the algorithm converge. Hence, to handle black-box constraints, we have selected the Scalable Constrained Bayesian Optimization (SCBO)~\cite{scbo}, based on the Trust Region Bayesian Optimization algorithm (TuRBO)~\cite{turbo} to tackle high dimensional search spaces and expensive objective function.

Like TuRBO, SCBO is based on Bayesian Optimization~\cite{garnett}, employing a probabilistic surrogate model, usually a Gaussian Process, to approximate the objective function and constraints. Different combinations of Matèrn kernel ($1/2$, $3/2$ and $5/2$) were tested to model different landscape of the covariance function. We kept the Matèrn$5/2$ since it was the most suitable one. Additive models were discarded as they generated numerical instability, particularly for the constraints' models.
SCBO and TuRBO instantiate trust regions to restrict the search within regions where the surrogate model is considered reliable, helping to focus on promising areas. These trust regions increase the scalability of SCBO in high dimensional search spaces.

The parallelization of SCBO is necessary since the early stopping criterion generates a high variability in the computation time of SNNs. The asynchronous strategy consists in maximizing the workload of processes by keeping a continuous flow of solutions. Thus, Thompson sampling was used as an acquisition function, since it can be parallelized asynchronously~\cite{async_ts}.

\subsection{Search spaces definition}

We optimized up to 19 HPs, by designing 4 experiments, and so 4 different search spaces, to study behaviors of HPO on SNNs trained with STDP and SLAYER, on digital Poisson converted MNIST and the neuromorphic DVS128 Gesture datasets.

The search spaces and best HPs found by the optimization are summed up in appendix \ref{ap:exps} and categorized into five HPs groups, from G1 to G5. In all 4 experiments, we optimized HPs of the neuron's model (G1), training algorithm (G2), architecture (G3), decoder or loss (G4) and training pipeline (G5).

HPs of the encoders are fixed. Indeed, preliminary experiments optimizing the encoding method (Poisson Encoded VS Time-To-First-Spike) and the presentation time ($T$) ranging from 10 to 300ms, showed a very high dependency of the accuracy on these HPs. This strong correlation, greatly affected the impact of more subtle HPs. These encoding HPs could be optimized in a multi-objective context, where number of spikes and accuracy are concurrent objectives. Moreover, $T$ has also a great impact on the training time, memory complexity and latency of the network. The greater $T$, the more expensive are the computations and memory usage. The architectural HPs $Exc$ and $Inh$ are the strength, i.e., fixed weights, between the excitatory and inhibitory layer. These HPs model Winner-Takes-All (WTA) or lateral inhibition mechanisms. Most of the HPs of the neurons' model are optimized for both layers. Traces time constant $\tau_{trace}$ are fixed, as well as both reset potentials $V_{reset}$. Weight normalization is described in~\cite{diehl}, and consists in enforcing the sum of synaptic weights from a neuron to be equal to a certain value. So the upper bound is fixed according to enforcing all weights from inputs to excitatory neurons, to be equal to 1.

The architecture of experiment 1 is the one found in~\cite{diehl}. The architecture of experiment 2, is slightly different and instantiates a distance-based soft lateral-inhibition. Neurons that are close together strongly inhibit each other, and conversely~\cite{hazan, iyer}. In experiment 2, because DVS128 Gesture is a harder dataset, Support Vector Machine (SVM) and Log Regression can, if selected, decode the total number of outputs spikes for each output neuron. These two decoders are often used in the literature, even if it is not considered as a fully spiking solution. This experiment also introduces the “Reset interval” HP~\cite{iyer}, which prevents too high temporal dependency on previous frames. During the presentation of a DVS data, neurons' parameters are periodically reset to their initial state.

Training of networks, in experiments 1 and 2, are under two stopping criterions and their corresponding constraints on the excitatory and inhibitory layers. For experiment 1, SNNs are expected to output at least $\alpha=5$ spikes at the excitatory outputs, and at least $\alpha=1$ spike at the inhibitory output for at least $90\%$ ($\beta=10\%$) of the training dataset.  In experiment 2, because of the higher complexity of DVS128 Gesture, we allow a higher flexibility, $\alpha=1$ and $\beta=30\%$ for both stopping criterions.

Concerning experiments 3 and 4 based on SLAYER, both used architecture can be found in \cite{slayer}. Instead of SRM neurons used in the original paper, in this work we have decided to use LIF neurons with adaptive threshold. In experiments 3, the parameters of the stopping criterions are set to $\alpha=3$ and $\beta=5\%$. In experiments 4, $\alpha=1$ and $\beta=30\%$. We allow a higher flexibility as the number of samples within DVS128 Gesture is much lower than MNIST.

Boundaries of the HPs heavily implied in the spiking activity, such as, $V_{\text{th}}$(neuron's threshold), $V_{\text{rest}}$(neuron's reset potential), $\tau$(leakage) or $t_{\text{ref}}$ (refractory period) were defined by uniformly sampling random combinations of HPs on a reduced subset of MNIST and DVS Gesture, until some networks present a minimal spiking activity. However, we did not try to prevent silent networks, so to define a more general search space. The HPs having an impact on the memory complexity (e.g., number of neurons, number of kernels, kernel size or batch size), were set according to the available memory on a single GPU. The number of epochs was set according to computation time of a single solution and the budget of one experiment.
To bias the initial combinations of HPs for SCBO, these are sampled differently according to their additive or multiplicative effect, or to bias the sampling toward known suitable solutions. For instance, the leakage $\tau$ of a neuron, because of its multiplicative effect, is sampled according to a log-uniform distribution. When values closer to the upper bound need to be sampled with a higher probability, the log-uniform distribution is reversed (R-LogUniform).

\begin{table*}[]
    \centering
    \caption{Summary of all 4 experiments}\label{tab:my_label}
    \begin{tabular}{c|c|c|c|c|c|c|c}
         \# & Dataset & Data shape & Architecture & Training & Simulator & HPs & Constraints\\
         \hline
         1 & MNIST & $B.100.1.28.28$ & SOM & STDP + Decoder & BindsNET & 18 & 2\\
         2 & DVS128 Gesture & $B.100.2.128.128$ & SOM & STDP + Decoder & BindsNET & 19  & 2\\
         \hline
         3 & MNIST & $B.25.1.28.28$ & CSNN~\cite{slayer} & SLAYER & LAVA-DL & 17  & 1\\
         4 & DVS128 Gesture & $B.100.2.128.128$ & CSNN~\cite{slayer} & SLAYER & LAVA-DL & 17  & 1\\
    \end{tabular}
\end{table*}

\subsection{Simulators and datasets}

To overcome the hardware bottleneck, one can mimic behaviors of SNNs by using simulators. Depending on the use case, the selection of the most suitable simulator has to be made carefully. Thus, we have selected Bindsnet~\cite{bindsnet} since it handles biologically inspired training algorithms (2-factor STDP, 3-factor STDP), it is based on PyTorch, so it can be accelerated on GPU. Bindsnet also implements different neuron models, IF, LIF or Izhekevich are available. One can also find different encoding methods such as Poisson encoder or Rank encoding, and decoding methods such as Average Spikes~\cite{voting}, Max Spikes~\cite{shahsavari} or $n$-gram~\cite{hazan}. Bindsnet is used for experiments in Table \ref{tab:exp1} and \ref{tab:exp2}.
We also used a second simulator, Lava-DL, as it is also based on PyTorch, and it implements the SLAYER~\cite{slayer} surrogate gradient supervised training algorithm. Experiments in Table \ref{tab:exp3} and \ref{tab:exp4} were simulated with LAVA-DL.

Performances of SNN are often assessed on standard ANN classification datasets, such as, Poisson encoded MNIST~\cite{zhou, kulkarni, diehl, hazan, saunders_scnn}. But these benchmarks should be considered a proof-of-concept for SNNs, and spiking analog dataset should be preferred to test SNNs performances~\cite{pfeiffer, malcolm}. Therefore, we have selected two benchmarks: Poisson encoded MNIST~\cite{mnist} and DVS128 Gesture~\cite{dvsgesture}.

In this work, MNIST was encoded within 100 frames, the shape of the data is the following: $B.T.C.H.W$, for batch size, frames, channels, height, and width ($B.100.1.28.28$). The number of frames is $3.5$ time lower than in~\cite{diehl}. No other transformation was made, such as denoising or centering. For experiment 3 based on SLAYER, $T$ was set to 25~\cite{slayer}.

Concerning DVS128 Gesture, spikes were accumulated within 100 frames, overlapping spikes during this process are considered as a single spike. Data was denoised using a 5000ms temporal neighborhood. Both \texttt{ON} and \texttt{OFF} channels are considered, so the shape is $B.100.2.128.128$. The Tonic~\cite{tonic} Python package was used to process the data. The higher pixel and temporal resolutions of DVS128 Gesture is a real challenge as it involves higher topology, memory, and computation complexities.

Both datasets were divided into training, validation, and testing datasets of respective sizes 48000, 12000 and 10000 for MNIST. DVS128 Gesture is divided into datasets of sizes 862, 215 and 264. The optimized accuracy is the one obtained on the validation dataset, and final results of the best solution found are assessed on the testing datasets.

\subsection{Hardware and software specifications}

Long-run experiments were carried out on the GPU partition of the Jean Zay supercomputer. Each experiment lasted for 100h, 15 Nvidia Tesla V100 with 32Gb of RAM were dedicated to the computation of SNNs, one additional GPU was used for the computations of SCBO (e.g., Gaussian processes). A single experiment represents a total of 1600 GPU hours. 
The 16 GPUs are grouped by clusters of 4, containing 2 Intel Cascade Lake 6248 processors of 20 cores each, a cluster cumulates a total of 160Gb of RAM.

The experiments were parallelized using OpenMPI interfaced by the python library \texttt{mpi4py}. BindsNET is fully based on \texttt{PyTorch}, while LAVA-DL also compiles custom CUDA code, both can easily be run onto Nvidia GPUs. SCBO was implemented and instantiated using \texttt{Zellij}\footnote{https://github.com/ThomasFirmin/zellij} and \texttt{BoTorch}\cite{botorch}.

\section{Computational results on large-scale experiments}\label{sec:results}

\subsection{Analysis of the HPO process}

\begin{figure*}[ht]
     \centering
     \begin{subfigure}[b]{0.49\textwidth}
         \centering
         \includegraphics[trim={0 0 0 0.8cm},clip,width=\textwidth]{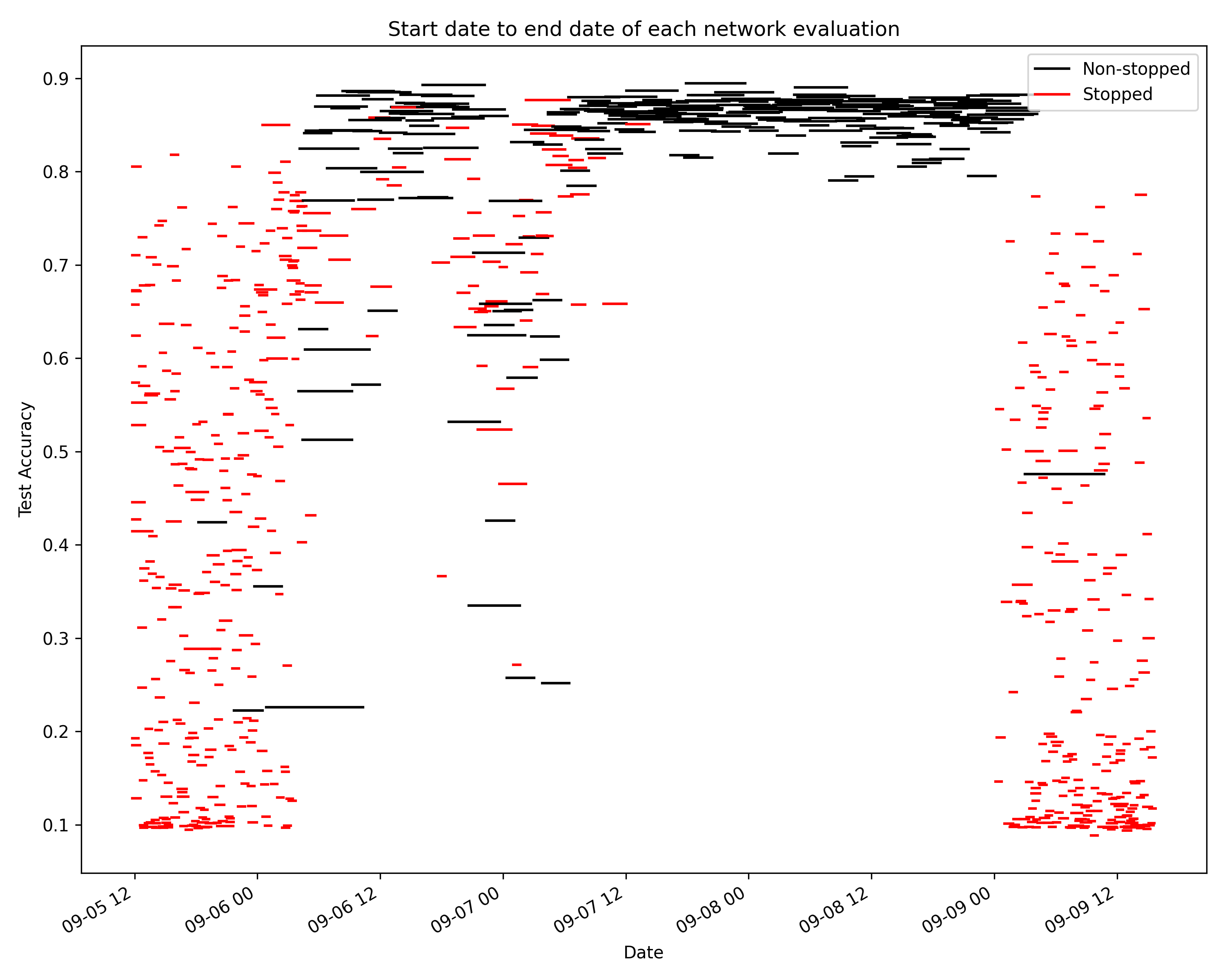}
         \caption{Experiment 1}\label{fig:sedate_mnist_100}
     \end{subfigure}
     \begin{subfigure}[b]{0.49\textwidth}
         \centering
         \includegraphics[trim={0 0 0 0.8cm},clip,width=\textwidth]{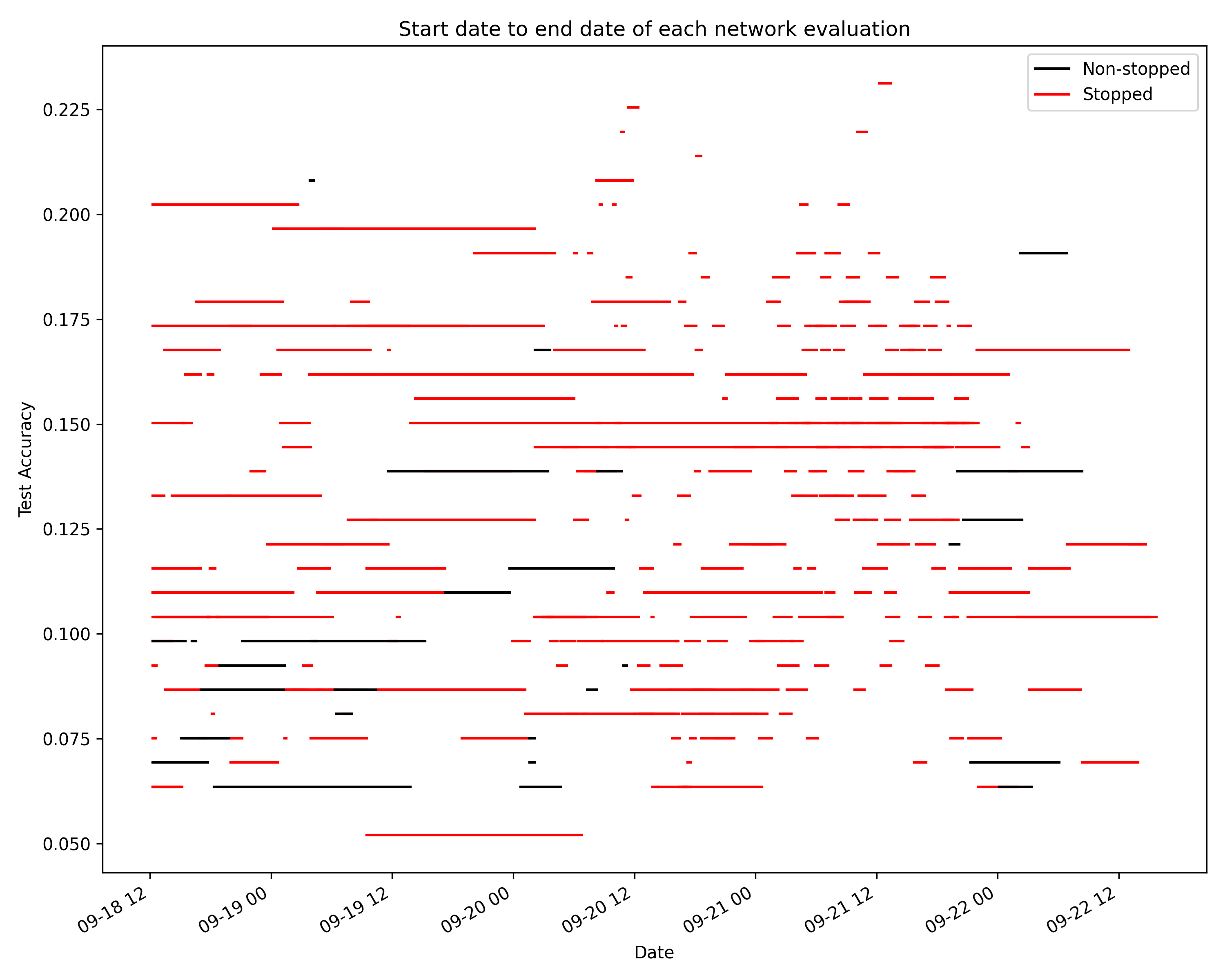}
         \caption{Experiment 2}\label{fig:dvs_stdp_date}
     \end{subfigure}
     \begin{subfigure}[b]{0.49\textwidth}
         \centering
         \includegraphics[trim={0 0 0 0.8cm},clip,width=\textwidth]{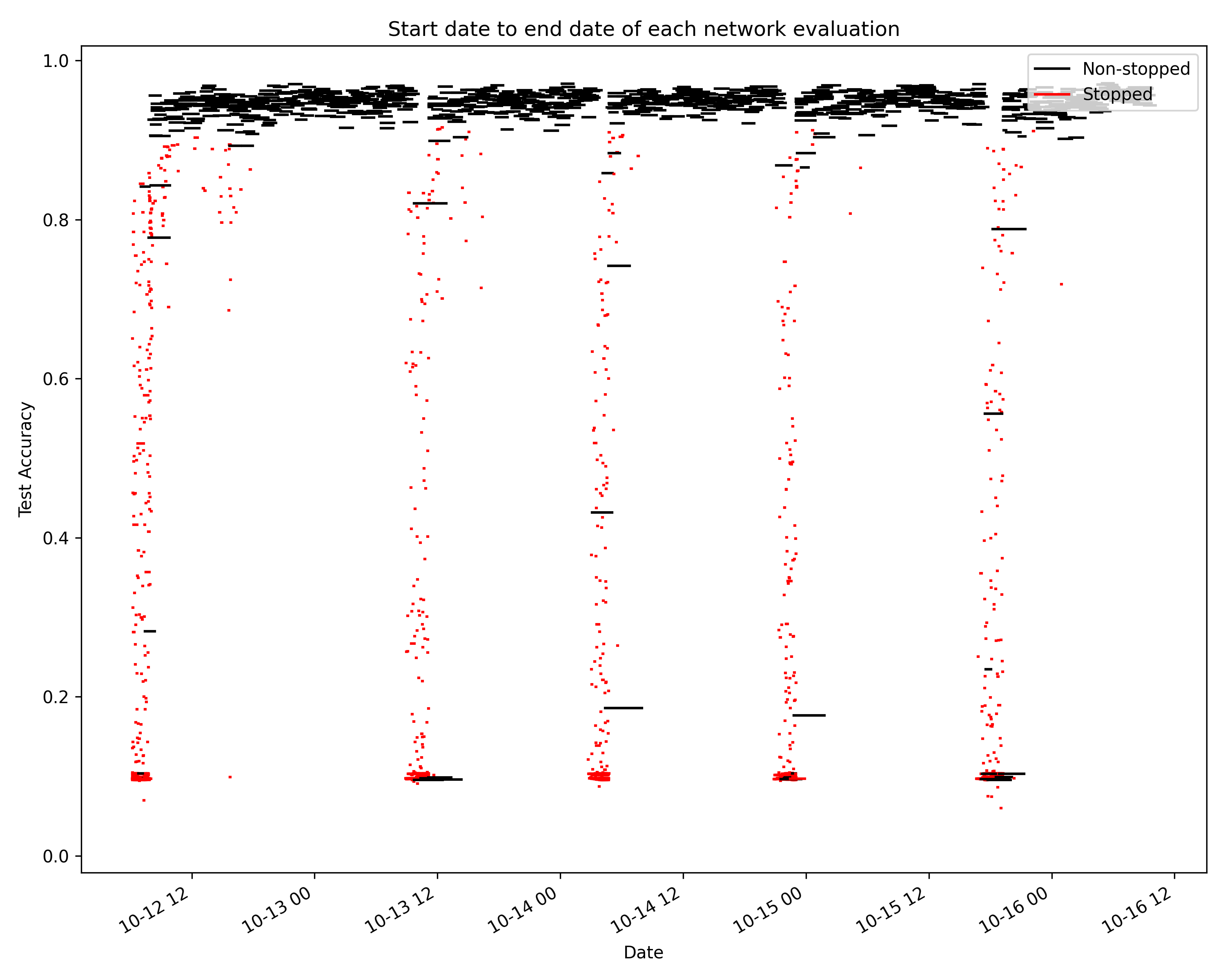}
         \caption{Experiment 3}\label{fig:mnist_25_date}
     \end{subfigure}
     \begin{subfigure}[b]{0.49\textwidth}
         \centering
         \includegraphics[trim={0 0 0 0.8cm},clip,width=\textwidth]{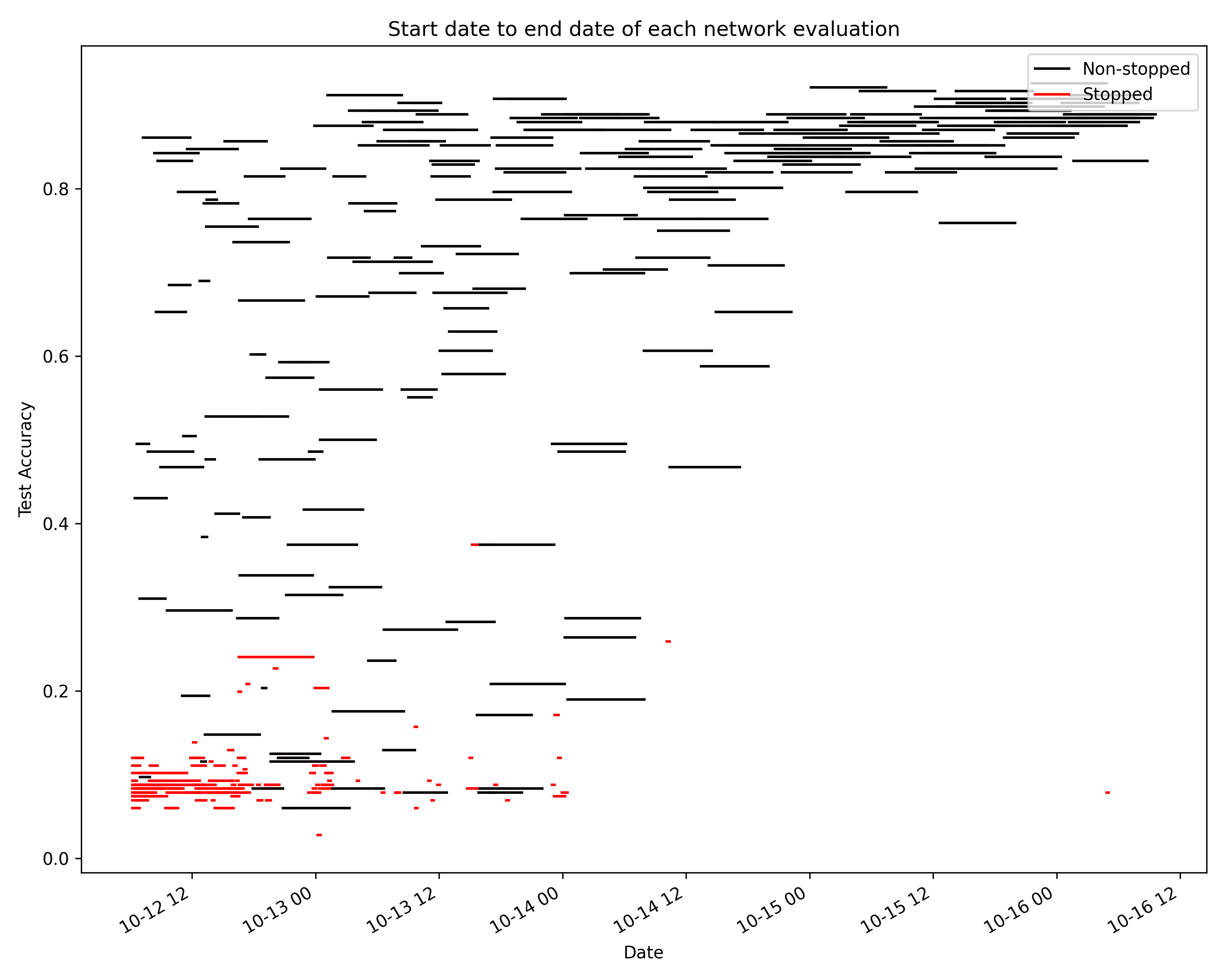}
         \caption{Experiment 4}\label{fig:dvs_slayer_date}
     \end{subfigure}
    \caption{Computation start date to end date compared to accuracy of each SNN.}
    \label{fig:exp_dates}
\end{figure*}

In the following lines, an analysis of the impact of silent-networks on SCBO is made.
In Figure \ref{fig:sedate_mnist_100} to \ref{fig:dvs_slayer_date}, a single horizontal line corresponds to the starting and ending dates of the evaluation of a single SNN. This representation emphasizes the ability of SCBO to detect silent-network and focus on fully trained SNNs with high accuracies. The observed drops in accuracy and computation time, for experience 1 and 3, are explained by the reset of the trust region once it shrank to its limits. Then, new random points are sampled, computed, added to the list of existing ones, and SCBO restarts the process.

\begin{figure*}[]
     \centering
     \begin{subfigure}[b]{0.49\textwidth}
         \centering
         \includegraphics[trim={{0.2\wd0} 0 {0.2\wd0} 0},clip,width=\textwidth]{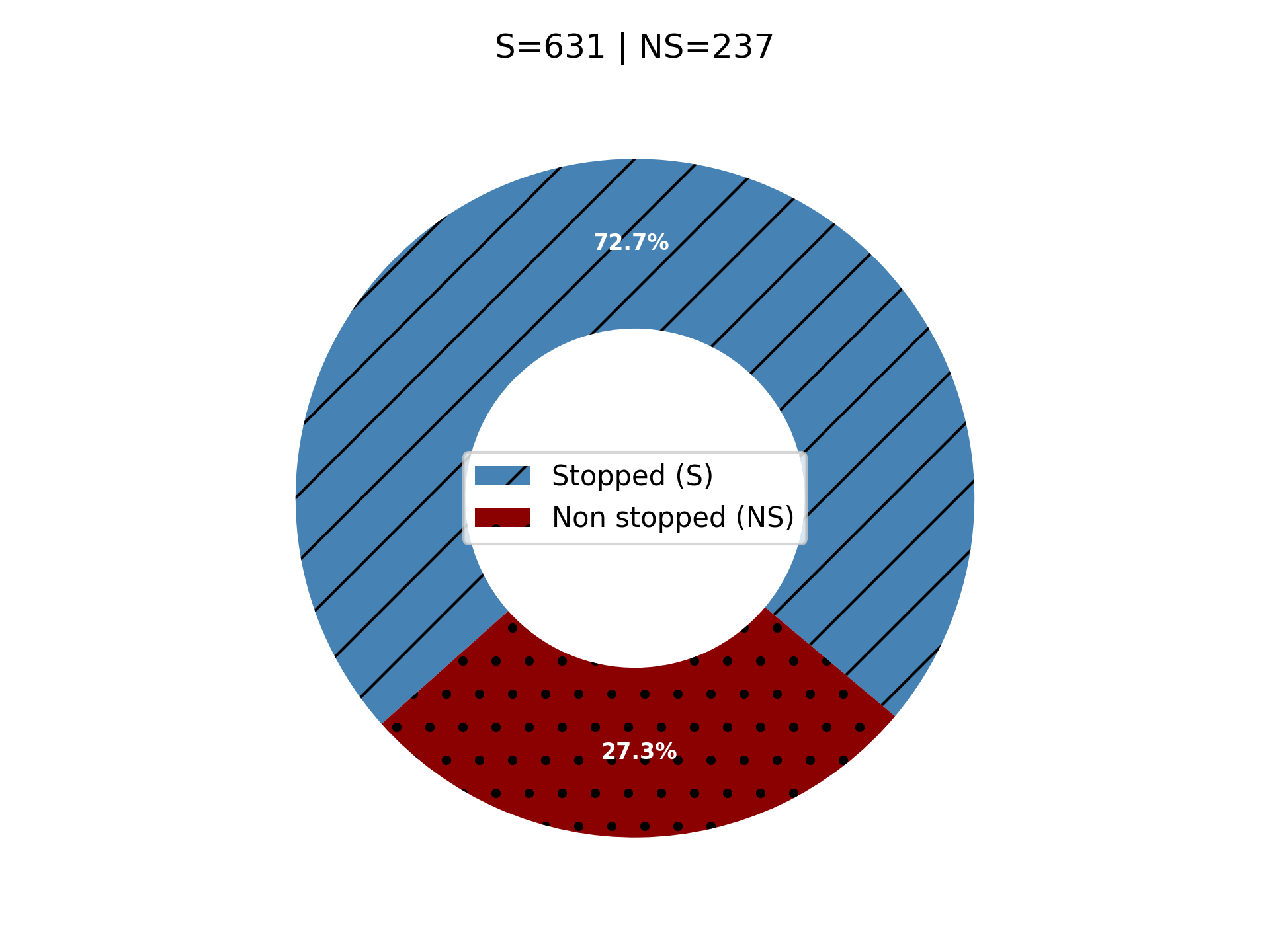}
         \caption{Proportion of early stopped network.}
     \end{subfigure}
     \begin{subfigure}[b]{0.49\textwidth}
         \centering
         \includegraphics[trim={{0.2\wd0} 0 {0.2\wd0} 0},clip,width=\textwidth]{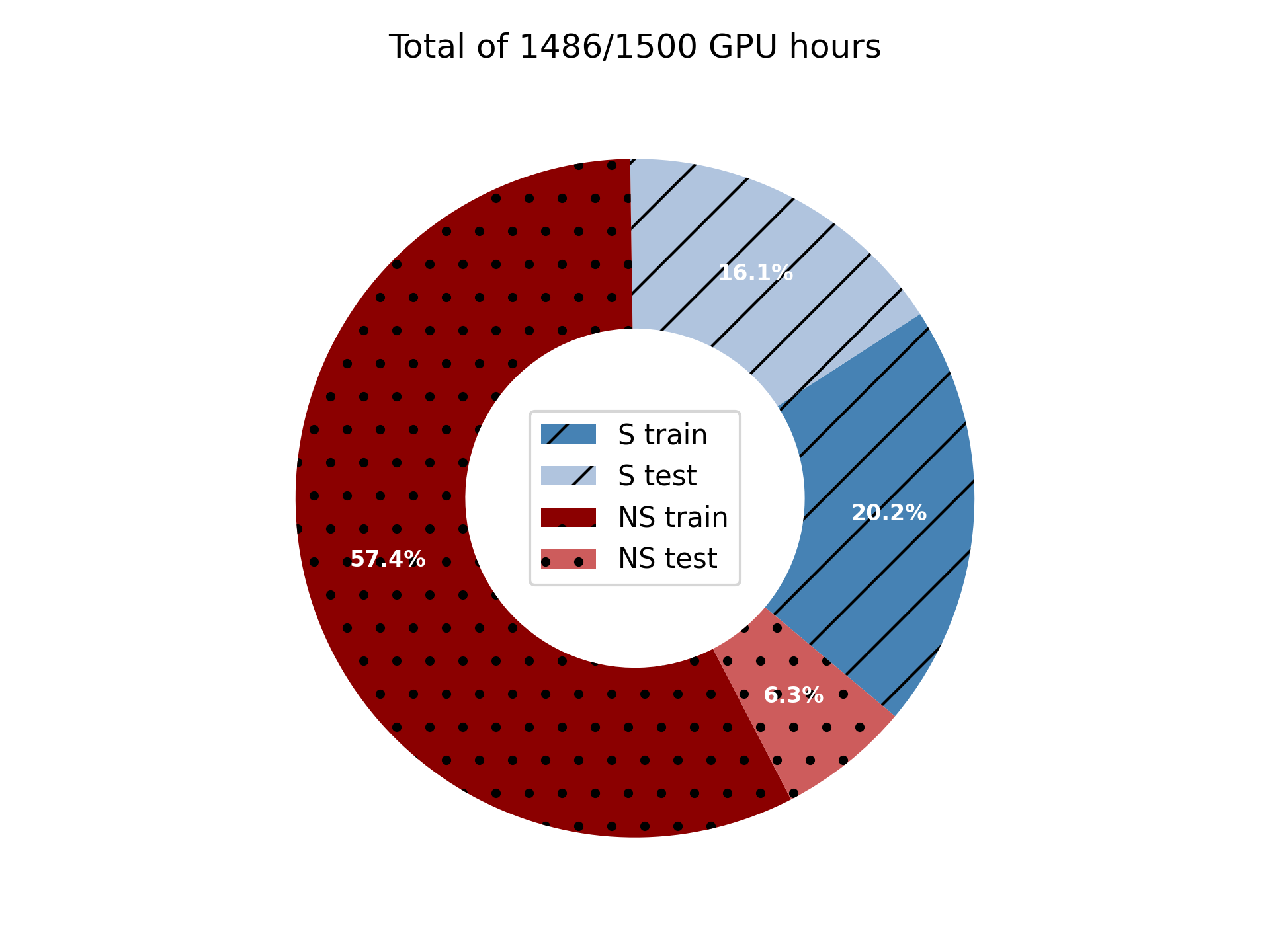}
         \caption{Cumulated training and validation time.}
     \end{subfigure}
    \caption{Stopped and non-stopped networks during experiment 1.}
    \label{fig:exp1_prop}
\end{figure*}

During experiment 1, one can see, in Figure \ref{fig:exp1_prop}, while almost $73\%$ of the evaluated networks were stopped, silent networks only consumed about $36\%$ of the 1500GPU hours. So, for experiment 1, the early stopping criterion and constraints worked. Indeed, Figure \ref{fig:sedate_mnist_100} emphasizes the focus of SCBO on non-silent networks, resulting in high validation accuracies.

Concerning experiment 2, the HPO failed to focus on feasible solutions. Only 321 networks were computed, and among them about $92\%$ were stopped, and conversely to other experiments, about $86\%$ of the time budget was spent on computing silent networks. These can be explained, by the expensive computation time, which can reach up to 50 hours, for a shallow network of 2 layers on BindsNET. The SOM architecture might not be suited for such a task, as it involves about $10^7$ parameters for only two layers.

\begin{figure*}[]
     \centering
     \begin{subfigure}[b]{0.49\textwidth}
         \centering
         \includegraphics[trim={{0.2\wd0} 0 {0.2\wd0} 0},clip,width=\textwidth]{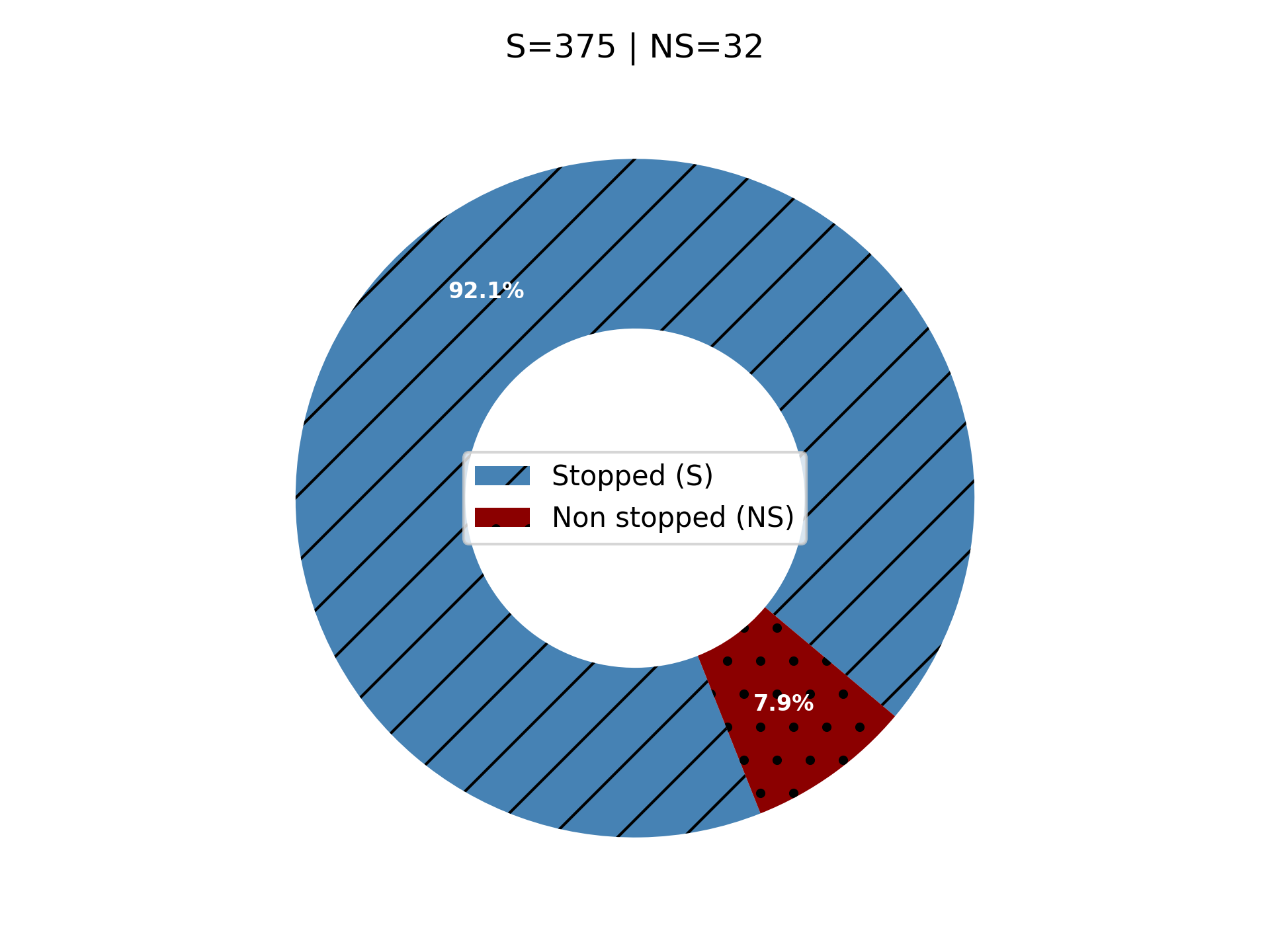}
         \caption{Proportion of early stopped network.}
     \end{subfigure}
     \begin{subfigure}[b]{0.49\textwidth}
         \centering
         \includegraphics[trim={{0.2\wd0} 0 {0.2\wd0} 0},clip,width=\textwidth]{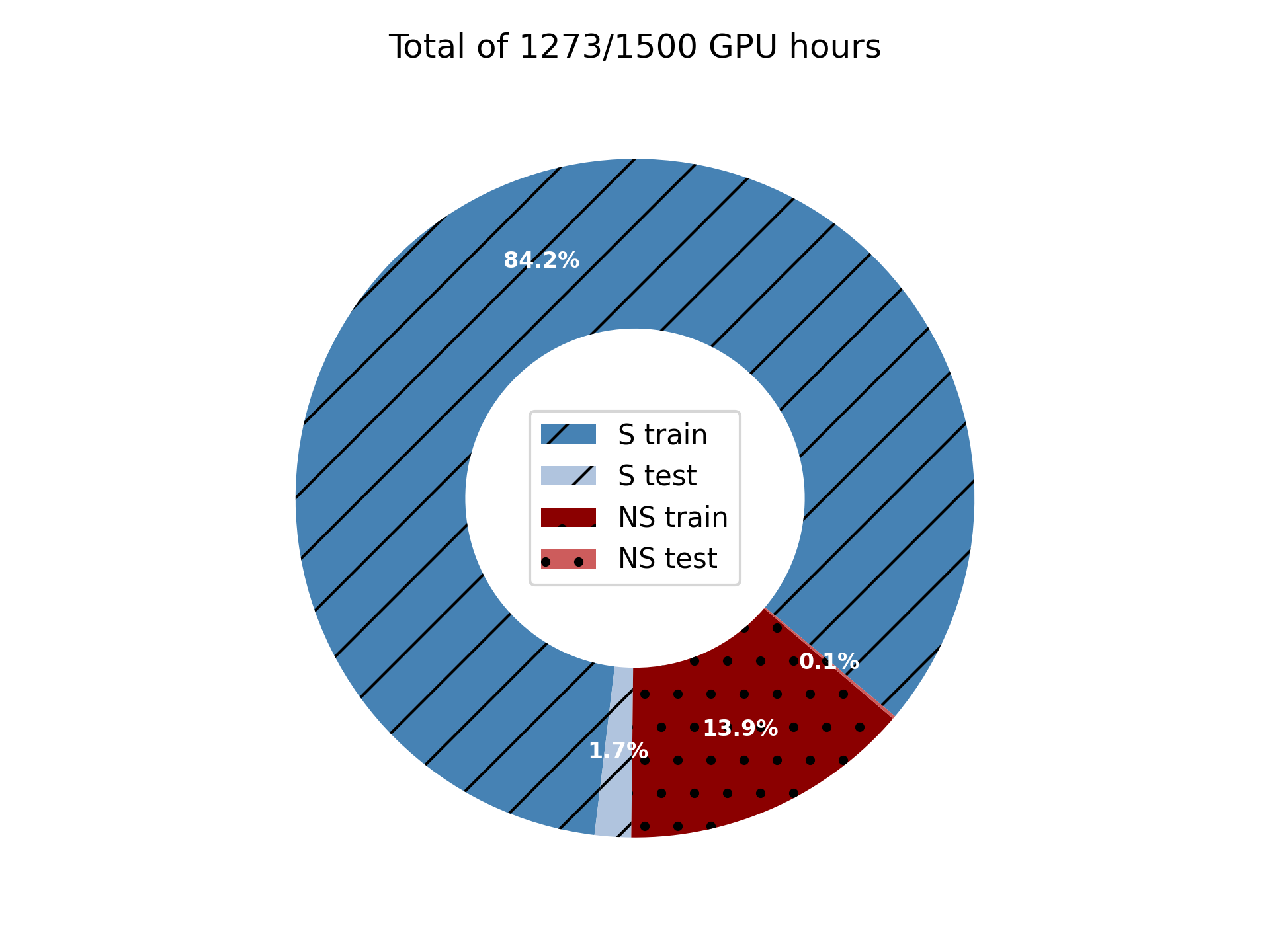}
         \caption{Cumulated training and validation time.}
     \end{subfigure}
    \caption{Stopped and non-stopped networks during Experiment 2.}
    \label{fig:exp2_prop}
\end{figure*}

\begin{figure*}[]
     \centering
     \begin{subfigure}[b]{0.49\textwidth}
         \centering
         \includegraphics[trim={{0.2\wd0} 0 {0.2\wd0} 0},clip,width=\textwidth]{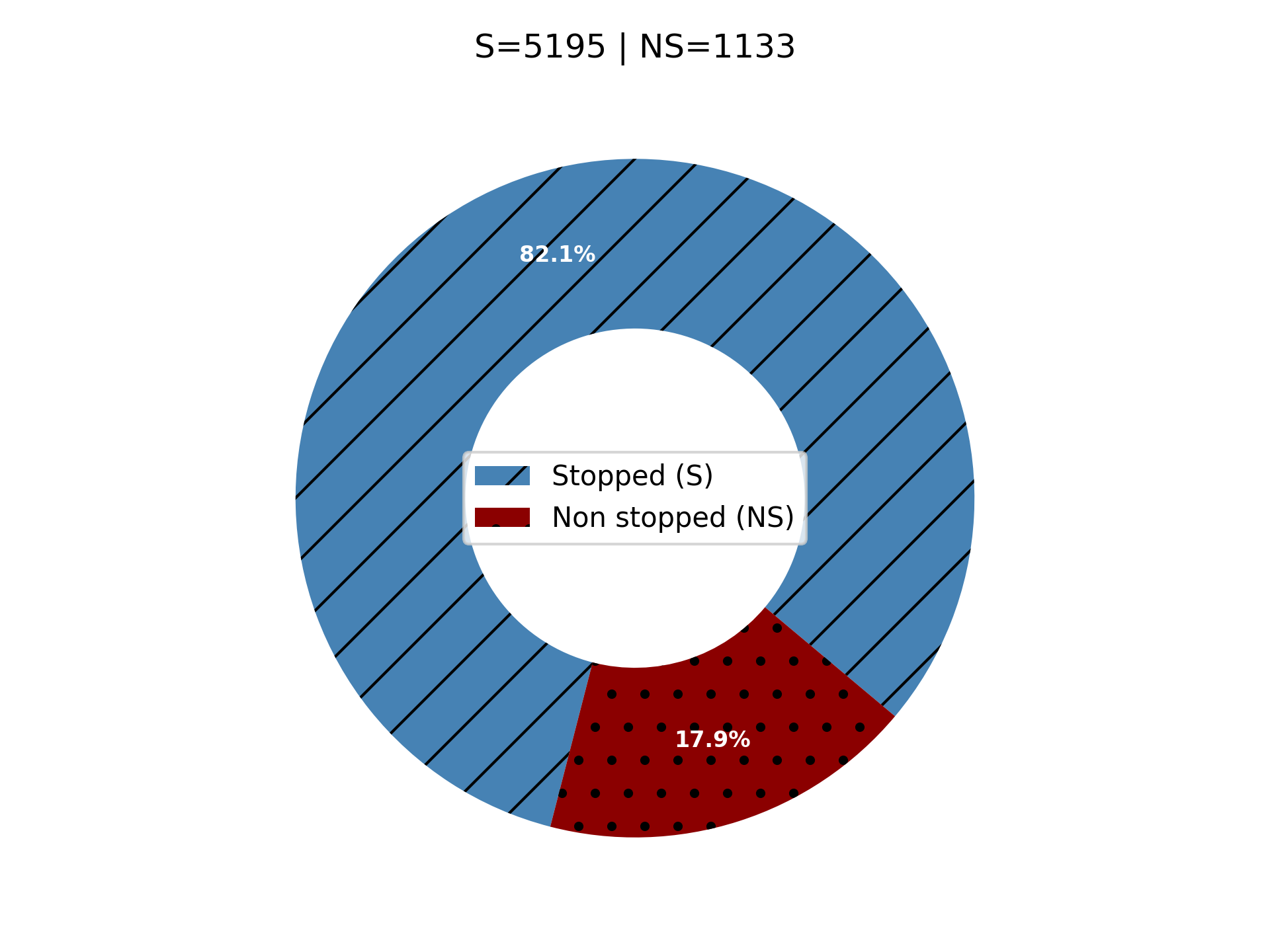}
         \caption{Proportion of early stopped network.}
     \end{subfigure}
     \begin{subfigure}[b]{0.49\textwidth}
         \centering
         \includegraphics[trim={{0.2\wd0} 0 {0.2\wd0} 0},clip,width=\textwidth]{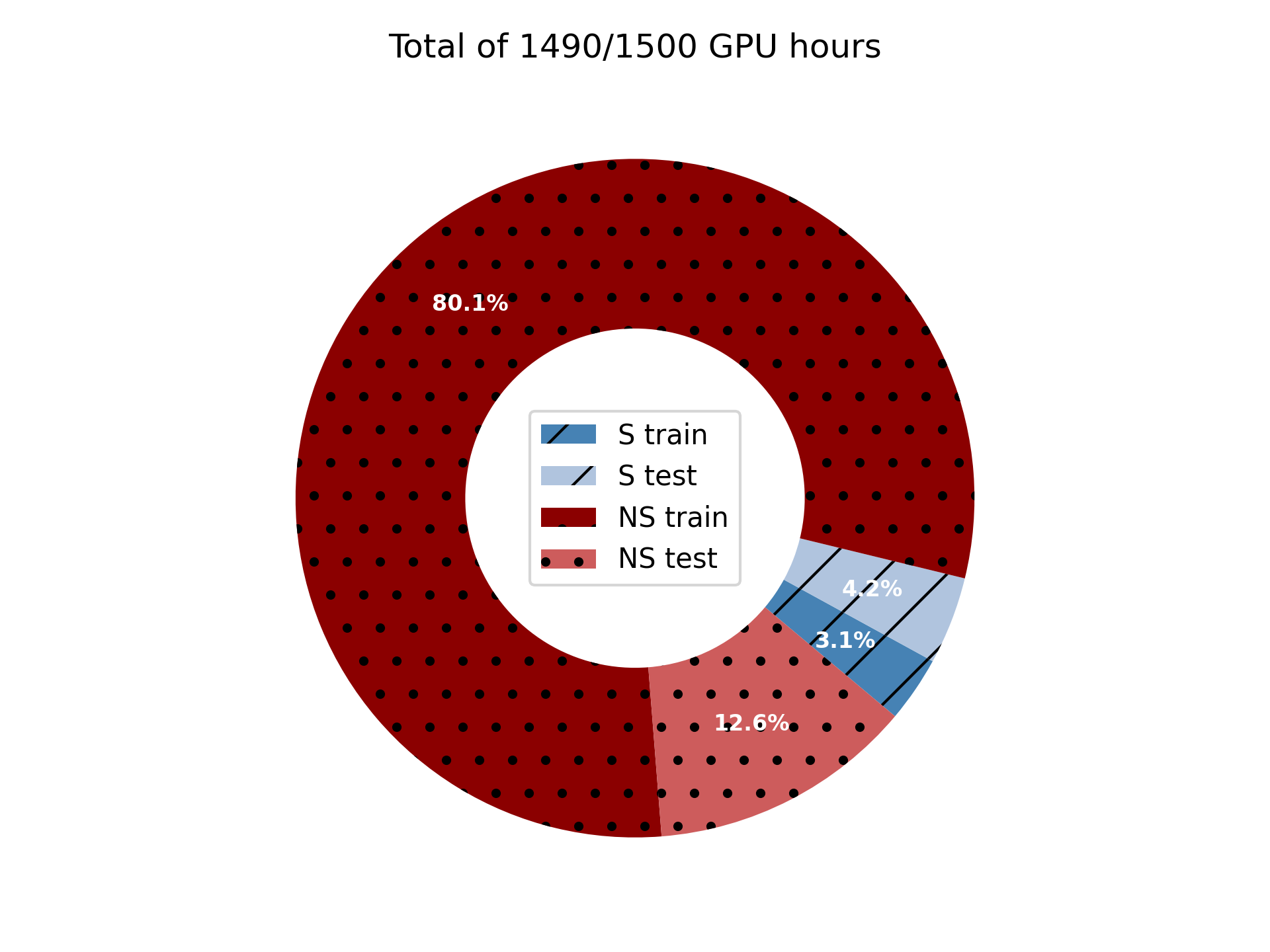}
         \caption{Cumulated training and validation time.}
     \end{subfigure}
    \caption{Stopped and non-stopped networks during experiment 3.}
    \label{fig:exp3_prop}
\end{figure*}

\begin{figure*}[]
     \centering
     \begin{subfigure}[b]{0.49\textwidth}
         \centering
         \includegraphics[trim={{0.2\wd0} 0 {0.2\wd0} 0},clip,width=\textwidth]{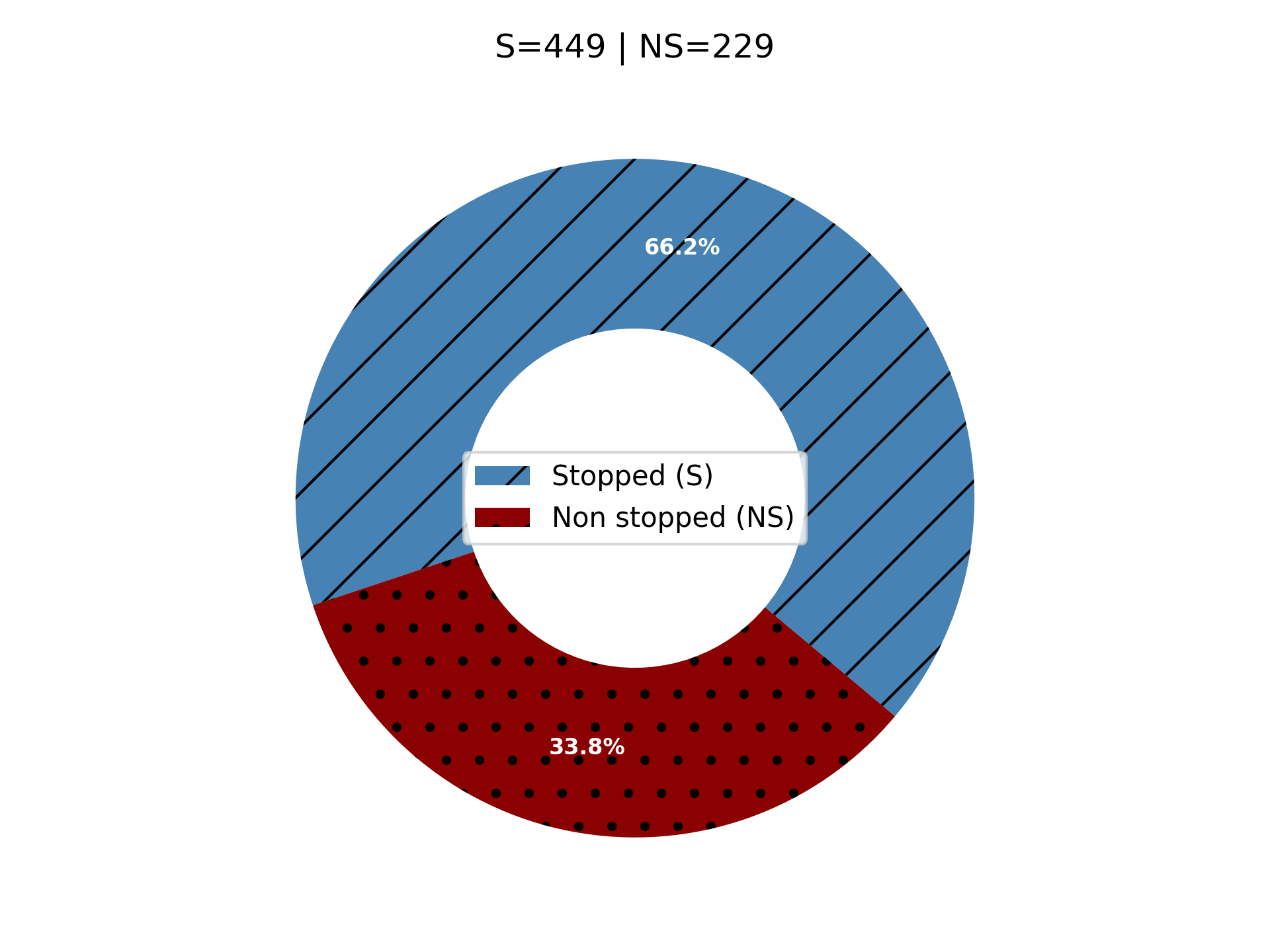}
         \caption{Proportion of early stopped network.}
     \end{subfigure}
     \begin{subfigure}[b]{0.49\textwidth}
         \centering
         \includegraphics[trim={{0.2\wd0} 0 {0.2\wd0} 0},clip,width=\textwidth]{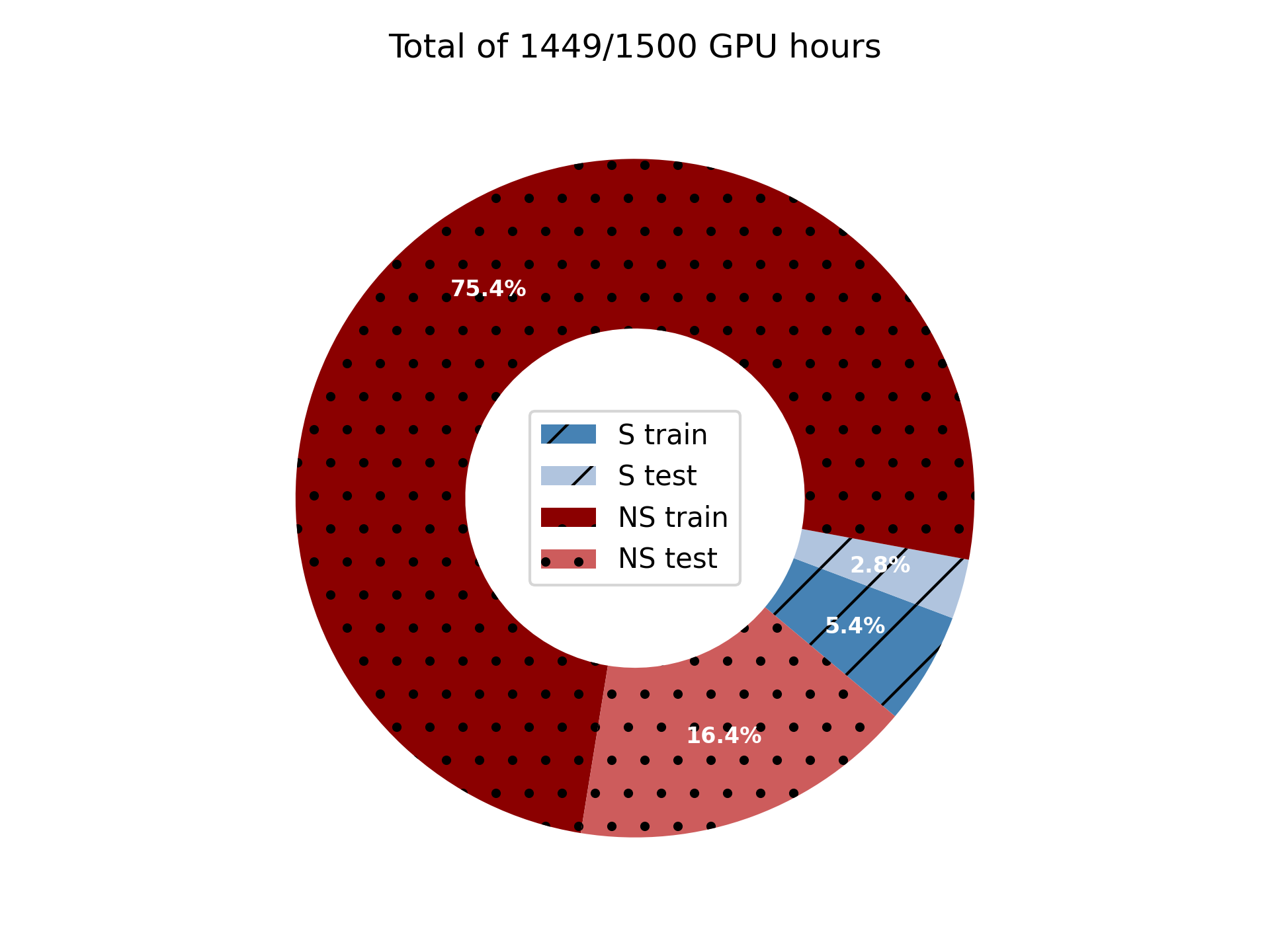}
         \caption{Cumulated training and validation time.}
     \end{subfigure}
    \caption{Stopped and non-stopped networks during experiment 4.}
    \label{fig:exp4_prop}
\end{figure*}

Similar behaviors to experiment 1 are observed in experiments 3 and 4 using SLAYER. Because LAVA-DL appears to be faster than BindsNET, more networks can be computed during the optimization process. In Figure \ref{fig:exp3_prop} and \ref{fig:exp4_prop}, by considering silent networks, the acceleration is even more emphasized compared to experiments using BindsNET. In experiment 3, $82\%$ of the networks were stopped, but only consumed $7\%$ of the total GPU hours. Because DVS128 Gesture is more time-consuming, fewer networks were computed in experiments 4, so $66\%$ of the networks were stopped representing about $8\%$ of the total GPU hours. In experiment 3, SCBO restarted 5 times, while in experiment 4, no restart happened. But in both cases and for every restart, SCBO was able to focus on high accuracies and non-silent networks.

In experiment 1, 3 and 4, despite more general search spaces with higher dimensionalities and wider bounds, we were able to find suitable solutions with good accuracies by considering silent-networks as a part of the HPO process. SCBO was able to scale in dimension, and by modeling black-box constraints allowed to prevent sampling silent-networks.
An interesting point emphasized by all 4 experiments and shown in Figures \ref{fig:sedate_mnist_100}, \ref{fig:mnist_25_date} and \ref{fig:dvs_slayer_date}, is that early stopped networks, visible by a red horizontal line, can quickly have acceptable accuracies. This confirms that multi-fidelity HPO can be used on SNNs.

\subsection{Analysis of the best solutions found}

The best solutions found for experiments 1, 3 and 4, were retrained on a larger number of epochs and then, the final classification on the testing dataset was made to assess solutions' performances. Concerning experiment 2 for which the best SNN has a $23\%$ of accuracy on the validation dataset. We decided to not intensively retrain this network since it was stopped during training, it has a high computation time and low performances. This emphasizes the difficulty of obtaining acceptable results, in a reasonable amount of time, with STDP on DVS128 Gesture. This can be explained by an unsuitable architecture or by the sensitivity of the HPs and the difficulty of defining a viable search space. Experiment 2 was made several times, with different configurations, by reducing the search space or modifying the preprocessing of input data, similar results were obtained. Comparatively, experiment 4, achieves better accuracies, with much fewer efforts and less computation time.

Best, HPs found and accuracies are summed up in appendix \ref{ap:exps}. In experiment 1, considering the optimized decoder (MaxSpike), the best testing accuracy is equal to $88.4\%$ similarly to AverageSpike. Considering other decoders, the maximum testing and validation accuracies are achieved by the SVM with respectively $92\%$ and $90.3\%$.
The 2-gram decoder has a maximum of $90.2\%$ on the testing dataset, which is the best accuracy among low complexity decoders. The values of the optimized HPs $Exc$ and $Inh$ are high, this indicates a convergence of the HPO toward a WTA mechanism with MaxSpike. Decoding outputs using spikes counts and complex machine learning methods gives, here, better results, at the expense of a higher complexity and are non neuromorphic solutions.
Compared to~\cite{diehl}, we were able to achieve similar performances considering the number of neurons, but with 3.5 times less exposition time, and so lower latency and spikes. In~\cite{diehl}, authors achieved $87.0\%$ with 400 neurons, $91.9\%$ with 1600 and $95.0\%$ with 6400, using a 350ms exposition time. However, our training dataset was smaller due to the additional testing set. Moreover, in our case, a single image is not passed multiple times to the network until reaching a minimum of 5 output spikes~\cite{diehl}. This condition was modeled by a black-box constraint on the outputs and was met for $97.34\%$ of the images during training.

In Experiment 3 since many networks with good accuracies were found, we have selected the one with a high accuracy and non-overfitted learning curves.
Compared to experiment 1, experiment 3 achieves a better accuracy $97.2\%$ with convolutions, pooling and about 2.5 times fewer parameters. The training time per epoch is cheaper, about 6 minutes for a single epoch, while the other architecture on BindsNET needs 138 minutes for one epoch. In this experiment, we were able to easily obtain better accuracy, with a sampling duration ($T$) 4 times lower than experiment 1.
Compared to experiment 2, in experiment 4, it is much easier to obtain good accuracies in less time, with more epochs and parameters. It is worth noticing a high  value, of about $40\%$ of neurons dropout, a regularization technique. We found a maximum testing accuracy of $88.6\%$. However, we faced overfitting during intensive training of the best solution found. This can be partly explained by the fact that the networks are trained on fewer data (862 samples), due to the train, validation, and test splits. 

\begin{table*}[H]
\centering
\caption{Experimental results}\label{tab:results1}
\begin{tabular}{l|ccc|ccc}
 & \multicolumn{3}{c|}{Experiment 1} & \multicolumn{3}{c}{Experiment 2} \\
\hline
Decoder & Testing & Validation & Training & Testing & Validation & Training \\
MaxSpike & $86.8\pm1.6$ & $86.3\pm1.7$ & $86.0\pm1.1$ & $\emptyset$ & $23.1$&$16.7$ \\
AverageSpike & $87.1\pm1.3$ & $86.6\pm1.5$ & $86.4\pm0.8$ & $\emptyset$ & $23.1$&$15.6$ \\
2-gram & $88.8\pm1.4$ & $88.2\pm1.7$ & $86.8\pm1.6$ & $\emptyset$ & $15.6$&$14.2$ \\
LogReg & $90.5\pm1.0$ & $89.9\pm1.1$ & $89.5\pm0.5$ & $\emptyset$ & $21.4$&$9.4$ \\
SVM & $90.8\pm1.2$ & $90.2\pm1.1$ & $90.1\pm0.4$ & $\emptyset$ & $18.5$&$9.7$ \\
\hline
Epochs (Retrained) & \multicolumn{3}{c|}{8} & \multicolumn{3}{c}{Not retrained} \\
Time/Epoch & \multicolumn{3}{c|}{138 mins} & \multicolumn{3}{c}{Stopped} \\
Parameters & \multicolumn{3}{c|}{1 895 266} &\multicolumn{3}{c}{11 508 864} \\
Trainable parameters & \multicolumn{3}{c|}{624 848} &\multicolumn{3}{c}{11 272 192} \\
\hline
\hline
& \multicolumn{3}{c|}{Experiment 3} & \multicolumn{3}{c}{Experiment 4} \\
\hline
Loss & Testing & Validation & Training & Testing & Validation & Training \\
Rate & $96.4\pm0.8$ & $96.1\pm0.8$ & $96.3\pm0.7$ & $85.2\pm3.4$ & $92.8\pm3.5$ & $99.4\pm0.6$\\
\hline
Epochs (Retrained) & \multicolumn{3}{c|}{25} &\multicolumn{3}{c}{100} \\
Time/Epoch & \multicolumn{3}{c|}{5.8 mins} &\multicolumn{3}{c}{29.3 mins} \\
Parameters & \multicolumn{3}{c|}{770 741} &\multicolumn{3}{c}{145 605 592} \\
Trainable parameters & \multicolumn{3}{c|}{770 733} &\multicolumn{3}{c}{145 605 580} \\
\end{tabular}
\end{table*}

\section{Conclusion and future works} \label{sec:con}

Hyperparameter optimization of SNNs is challenging. This problem is often solved by thinking of the SNN as a fully black box. The spiking activity of SNNs is generally considered within a multi-objectives context, where it is minimized so to obtain a better energy efficiency.

However, ignoring that a SNN needs a minimal spiking activity, is ignoring the silent network problem explained by mistuned HPs or architecture. Therefore, during the HPO, SNNs should be instead considered as a gray-box, where information about its behaviors is used to better sample solutions.

Usually, search spaces are low dimensional and strongly bounded to only good solutions. In this work, we have shown that we can define general and scalable high dimensional search spaces containing many silent networks while maintaining performances of the HPO process for expensive SNNs.
By leveraging infeasible solutions, we can increase the efficiency of the exploration of search spaces. This is done by a combination of a spike based early stopping criterion and its associated black-box constraints. The early stopping criterions prevent useless computation by interrupting the training when a certain proportion of the data did not output enough spikes.
The Scalable Constrained Bayesian Optimization algorithm ensures a scalability in high dimension and models the constraints to prevent sampling silent networks.
Our approach was generalized to the two most popular families of training algorithms known as plasticity rules and surrogate gradient. For both cases, experimental results emphasize the value of our strategy, which maintains good performances within a generalized high dimensional search space.
Moreover, because of the early stopping criterion, computation time of a single SNN is stochastic. This stochasticity is handled by the asynchronous parallelization of the optimization algorithm on a heterogeneous multi-nodes and multi-GPU Petascale architecture.

Future works will focus on multi-objectives optimization, where finding the frontier between low-spiking and silent networks is crucial. Hybridizing our methodology with multi-fidelity algorithms, such as BOHB, is also under consideration. Finally, the strategy could also be considered within a Neural Architecture Search (NAS) framework. Only a few works have tackled this problem applied to SNNs~\cite{nasnn1,nasnn2}. Indeed, designing feasible solutions is difficult. Hence, by being able to quickly determine whether a given architecture is silent or not, could improve the optimization process.

\section*{Acknowledgments}

Design of experiments presented in this paper were carried out using the Grid'5000 testbed, supported by a scientific interest group hosted by Inria and including CNRS, RENATER and several Universities as well as other organizations (see https://www.grid5000.fr).

This work was granted access to the HPC resources of IDRIS under the allocation 2023-AD011014347 made by GENCI.

This work has been supported by the University of Lille, the ANR-20-THIA-0014 program AI\_PhD\@Lille and the ANR PEPR AI and Numpex. It was also supported by IRCICA(CNRS and Univ. Lille USR-3380).

\section*{Data availability statement}

All code and experimental datasets are available on GitHub at \url{https://github.com/ThomasFirmin/hpo_snn}.

\section*{Author contribution}
Thomas Firmin: Conceptualization, Formal analysis, Investigation, Methodology, Software, Validation, Visualization, Writing - original draft. El-Ghazali Talbi and Pierre Boulet: Funding acquisition, Project administration, Supervision, Writing - review \& editing.

\clearpage

\clearpage
\onecolumn
\appendix
\section{Search spaces and optimized HPs}\label{ap:exps}
\begin{table}[h!]
\centering
\caption{Experiment 1, STDP trained SOM on MNIST}\label{tab:exp1}
\begin{tabular}{lllllll}
HP & Lower bound & Upper bound & Optimized & Sampler & Type & Group \\
\hline
\hline
$\lambda_-$ & 1e-4 & 1e-2 & 0.00084 & R-LogUniform & Continuous & G2 \\
$\lambda_+$ & 1e-4 & 1e-2 & 0.0088 & LogUniform & Continous & G2 \\
Map size & 20 & 2000 & 797 & Uniform & Discrete & G3 \\
Decoder & \multicolumn{2}{|c|}{Average, Max, 2-gram, 3-gram} & Max & Random choice & Categorical & G4 \\
Epochs & 1 & 3 & 2 & Uniform & Discrete & G5 \\
Weight Norm. & 78.4 & 784 & 123.38 & Uniform & Continuous & G5 \\
\multicolumn{7}{c}{Excitatory layer} \\
\hline
$V_{th}$ & -59 & 0 & -57.6 &Uniform & Continuous & G1 \\
$V_{rest}$ & -70 & -60 & -60.8 & Uniform & Continuous & G1 \\
$\tau$ & 5 & 5000 & 4166.8 & LogUniform & Continuous & G1 \\
$t_{ref}$ & 0 & 20 & 6 & Uniform & Discrete & G1 \\
$\theta_+$ & 0.001 & 0.5 & 0.044 & LogUniform & Continuous & G1 \\
$\tau_{\theta}$ & 1e6 & 1e7 & 2041798 & LogUniform & Continuous & G1 \\
$Exc$ & 0.5 & 500 & 356.9 & LogUniform & Continuous & G3 \\
\multicolumn{7}{c}{Inhibitory layer} \\
\hline
$V_{th}$ & -40 & 0 & -22.8 & Uniform & Continuous & G1 \\
$V_{rest}$ & -60 & -45 & -51 & Uniform & Continuous & G1 \\
$\tau$ & 5 & 5000 & 1516.15 & LogUniform & Continuous & G1 \\
$t_{ref}$ & 0 & 20 & 19 & Uniform & Discrete & G1 \\
$Inh$ & 0.5 & 500 & 433.6 & LogUniform & Continuous & G3 \\
\hline
\hline
\multicolumn{7}{c}{Fixed}\\
\hline
$V_{reset}$ (Excitatory) & \multicolumn{2}{c}{-60} & & & Continuous & G1\\
$V_{reset}$ (Inhibitory) & \multicolumn{2}{c}{-45} & & & Continuous & G1\\
$\tau_{trace}$ (Inputs) & \multicolumn{2}{c}{20} & & & Continuous & G2\\
$\tau_{trace}$ (Excitatory) & \multicolumn{2}{c}{20} & & & Continuous & G2\\
\hline
\hline
\multicolumn{7}{c}{Early stopping}\\
HP & \multicolumn{2}{c}{Excitatory} & \multicolumn{2}{c}{Inhibitory} & Type & Group \\
\hline
$\alpha$ & \multicolumn{2}{c}{5} & \multicolumn{2}{c}{1} & Discrete & G5\\
$\beta$ & \multicolumn{2}{c}{0.1} & \multicolumn{2}{c}{0.1} & Continuous & G5\\
\end{tabular}
\end{table}

\begin{table}
\centering
\caption{Experiment 2, STDP trained SOM on DVS Gesture}\label{tab:exp2}
\begin{tabular}{lllllll}
HP & Lower bound & Upper bound & Optimized & Sampler & Type & Group \\
\hline
\hline
$\lambda_-$ & 1e-4 & 1e-2 & 0.00491 & R-LogUniform & Continuous & G2 \\
$\lambda_+$ & 1e-4 & 1e-2 & 0.00977 & LogUniform & Continous & G2 \\
Map size & 20 & 1000 & 344 & Uniform & Discrete & G3 \\
Decoder & \multicolumn{2}{|c|}{Average, Max, 2-gram, SVM, Log} & Max & Random choice & Categorical & G4 \\
Epochs & 1 & 3 & 1 & Uniform & Discrete & G5 \\
Weight Norm. & 3276.8 & 32768 & 17522.71 & Uniform & Continuous & G5 \\
Reset interval & 5 & 100 & 20 & Uniform & Discrete & G5\\
\multicolumn{7}{c}{Excitatory layer} \\
\hline
$V_{\text{th}}$ & -59 & 60 & -27.4 & Uniform & Continuous & G1 \\
$V_{\text{rest}}$ & -140 & -60 & -111.0 & Uniform & Continuous & G1 \\
$\tau$ & 5 & 5000 & 2475.24 & LogUniform & Continuous & G1 \\
$t_{\text{ref}}$ & 0 & 40 & 11 & Uniform & Discrete & G1 \\
$\theta_+$ & 0.1 & 1 & 0.972 & LogUniform & Continuous & G1 \\
$\tau_{\theta}$ & 1e6 & 1e7 & 16783940 & LogUniform & Continuous & G1 \\
$Exc$ & 1 & 500 & 22.32 & Uniform  & Continuous & G3 \\
\multicolumn{7}{c}{Inhibitory layer} \\
\hline
$V_{\text{th}}$ & -40 & 40 & 2.81 & Uniform & Continuous & G1 \\
$V_{\text{rest}}$ & -120 & -45 & -75.85 & Uniform & Continuous & G1 \\
$\tau$ & 5 & 5000 & 761.38 & LogUniform & Continuous & G1 \\
$t_{\text{ref}}$ & 0 & 40 & 36 & Uniform & Discrete & G1 \\
$Inh$ & 1 & 500 & 66.05 & Uniform & Continuous & G3 \\
\hline
\hline
\multicolumn{7}{c}{Fixed}\\
\hline
$V_{\text{reset}}$ (Excitatory) & \multicolumn{2}{c}{-60} & & & Continuous & G1\\
$V_{\text{reset}}$ (Inhibitory) & \multicolumn{2}{c}{-45} & & & Continuous & G1\\
$\tau_{\text{trace}}$ (Inputs) & \multicolumn{2}{c}{20} & & & Continuous & G2\\
$\tau_{\text{trace}}$ (Excitatory) & \multicolumn{2}{c}{20} & & & Continuous & G2\\
\hline
\hline
\multicolumn{7}{c}{Early stopping}\\
HP & \multicolumn{2}{c}{Excitatory} & \multicolumn{2}{c}{Inhibitory} & Type & Group \\
\hline
$\alpha$ & \multicolumn{2}{c}{1} & \multicolumn{2}{c}{1} & Discrete & G5\\
$\beta$ & \multicolumn{2}{c}{0.1} & \multicolumn{2}{c}{0.3} & Continuous & G5\\
\end{tabular}
\end{table}

\begin{table}
\centering
\caption{Experiment 3, CSNN trained by SLAYER on Poisson encoded MNIST}\label{tab:exp3}
\begin{tabular}{lllllll}
HP & Lower bound & Upper bound & Optimized & Sampler & Type & Group \\
\hline
\hline
$V_{\text{th}}$ & 0.4 & 4 & 1.354 & Uniform & Continuous & G1 \\
$\tau$ & 0.01 & 0.2 & 0.1255 & LogUniform & Continuous & G1 \\
$\theta_+$ & 0.001 & 0.25 & 0.1419 & LogUniform & Continuous & G1 \\
$\tau_{\theta}$ & 0.01 & 0.5 & 0.1735 & LogUniform & Continuous & G1 \\
$\tau_{\text{ref}}$ & 0.1 & 0.99 & 0.1031 & R-LogUniform & Continuous & G1 \\
$\tau_u$ & 0.1 & 0.99 & 0.5577 & Uniform & Continuous & G1 \\
$\lambda$ & 1e-3 & 1e-1 & 0.0022 & LogUniform & Continuous & G2 \\
$G$ & 0.5 & 1 & 0.7287 & LogUniform & Continuous & G2 \\
$\tau_g$ & 0.1 & 1 & 0.1734 & LogUniform & Continuous & G2 \\
Loss & \multicolumn{2}{|c|}{Rate, Max} & Rate & Random choice & Categorical & G4 \\
Epochs & 1 & 40 & 14 & Uniform & Discrete & G5 \\
Batch & 1 & 50 & 32 & R-LogUniform & Discrete & G5 \\
Neurons dropout & 0.01 & 0.90 & 0.0620 & LogUniform & Continuous & G5 \\
\multicolumn{7}{c}{1st Convolutional layer} \\
\hline
Kernels & 1 & 128 & 120 & Uniform & Discrete &  G3\\
Kernel size & 4 & 12 & 12 & Uniform & Discrete &  G3\\
\multicolumn{7}{c}{2nd Convolutional layer} \\
\hline
Kernels & 1 & 128 & 51 & Uniform & Discrete &  G3\\
Kernel size & 4 & 12 & 11 & Uniform & Discrete &  G3\\
\hline
\hline
\multicolumn{7}{c}{Fixed}\\
\hline
Padding (convolutions) & \multicolumn{3}{c}{0} & & Discrete & G3\\
Stride (convolutions) & \multicolumn{3}{c}{1} & & Discrete & G3\\
Dilation (convolutions) & \multicolumn{3}{c}{1} & & Discrete & G3\\
Pooling layers & \multicolumn{3}{c}{See~\cite{slayer}} & & & G3\\
\hline
\hline
\multicolumn{7}{c}{Early stopping}\\
HP & \multicolumn{4}{c}{Outputs} & Type & Group \\
\hline
$\alpha$ & \multicolumn{4}{c}{3} & Discrete & G5\\
$\beta$ & \multicolumn{4}{c}{0.05} & Continuous & G5\\
\end{tabular}
\end{table}

\begin{table}
\centering
\caption{Experiment 4, CSNN trained by SLAYER on DVS128 Gesture}\label{tab:exp4}
\begin{tabular}{lllllll}
HP & Lower bound & Upper bound & Optimized & Sampler & Type & Group \\
\hline
\hline
$V_{\text{th}}$ & 0.1 & 1 &  0.660 & Uniform & Continuous & G1 \\
$\tau$ & 0.01 & 0.9 &  0.8536 & LogUniform & Continuous & G1 \\
$\theta_+$ & 0.001 & 0.4 &  0.0957 & LogUniform & Continuous & G1 \\
$\tau_{\theta}$ & 0.01 & 0.5 &  0.0502 & LogUniform & Continuous & G1 \\
$\tau_{\text{ref}}$ & 0.1 & 0.99 &  0.1258  & R-LogUniform & Continuous & G1 \\
$\tau_u$ & 0.1 & 0.99 &  0.8847  & Uniform & Continuous & G1 \\
$\lambda$ & 1e-3 & 1e-1 &  0.0215 & LogUniform & Continuous & G2 \\
$G$ & 0.5 & 1 &  0.6827  & LogUniform & Continuous & G2 \\
$\tau_g$ & 0.1 & 1 &  0.1956 & LogUniform & Continuous & G2 \\
Loss & \multicolumn{2}{|c|}{Rate, Max} &  Rate  & Random choice & Categorical & G4 \\
Epochs & 1 & 15 &  14  & Uniform & Discrete & G5 \\
Batch & 1 & 5 &  4  & R-LogUniform & Discrete & G5 \\
Neurons dropout & 0.01 & 0.90 &  0.3949 & LogUniform & Continuous & G5 \\
\multicolumn{7}{c}{1st Convolutional layer} \\
\hline
Kernels & 1 & 36 &  15  & Uniform & Discrete &  G3\\
Kernel size & 4 & 48 &  20  & Uniform & Discrete &  G3\\
\multicolumn{7}{c}{2nd Convolutional layer} \\
\hline
Kernels & 1 & 128 &  29  & Uniform & Discrete &  G3\\
Kernel size & 4 & 48 &  12  & Uniform & Discrete &  G3\\
\hline
\hline
\multicolumn{7}{c}{Fixed}\\
\hline
Padding (convolutions) & \multicolumn{3}{c}{0} & & Discrete & G3\\
Stride (convolutions) & \multicolumn{3}{c}{1} & & Discrete & G3\\
Dilation (convolutions) & \multicolumn{3}{c}{1} & & Discrete & G3\\
Pooling layers & \multicolumn{5}{c}{See~\cite{slayer}} & G3\\
\hline
\hline
\multicolumn{7}{c}{Early stopping}\\
HP & \multicolumn{4}{c}{Outputs} & Type & Group \\
\hline
$\alpha$ & \multicolumn{4}{c}{1} & Discrete & G5\\
$\beta$ & \multicolumn{4}{c}{0.3} & Continuous & G5\\
\end{tabular}
\end{table}

\begin{table}
\centering
\caption{Hyperparameter description}\label{tab:hplist}
\begin{tabular}{lll}
HP & Description & Group \\
\hline
\hline
$V_{\text{th}}$ & Neuron's threshold. & G1 \\
$V_{\text{rest}}$ & Neuron's resting potential. & G1 \\
$V_{\text{reset}}$ & Neuron's reset potential. & G1\\
$\tau$ & Neuron's leakage time constant. & G1 \\
$\theta_+$ & Added value to the threshold, for threshold adaptation mechanism. & G1 \\
$\tau_{\theta}$ & Threshold adaptation time constant. Makes the threshold decrease through time. & G1 \\
$\tau_{\text{ref}}$ & Refractory period time constant for SLAYER. & G1 \\
$t_{\text{ref}}$ & Refractory period in time steps. & G1 \\
$\tau_u$ & Time constant for current decay. & G1 \\
$\lambda$ & Learning rate of gradient descent. & G2 \\
$\lambda_-$ & Learning rate of pre-synaptic spikes in STDP. & G2 \\
$\lambda_+$ & Learning rate of post-synaptic spikes in STDP. & G2 \\
$\tau_g$ & Relaxation of the spike function gradient. & G2 \\
$G$ & Controls the gradient flow across layers, and handles vanishing or exploding gradient. & G2 \\
$\tau_{\text{trace}}$ & Traces time constant of STDP. & G2\\
Map size & Number of output neurons in the Dielh \& Cook architecture. & G3 \\
Kernels & Number of filters in a convolution. &  G3\\
Kernel size & Size of filters in a convolution. &  G3\\
Padding & Padding HP for convolution. & G3\\
Stride & Stride HP for convolution. & G3\\
Dilation & Dilation HP for convolution. & G3\\
Loss & Loss function optimized by the backpropagation. & G4 \\
Decoder & Algorithm used to decode outputs of Hebbian-trained SNNs. & G4 \\
Epochs & Number of epochs. & G5 \\
Batch & Batch size. & G5 \\
Neurons dropout & Probability of removing neurons and their connections during training. & G5 \\
Weight Norm. & Weight normalization. Value at which the sum of neuron weights must be equal. & G5 \\
Reset interval & Time between two resets of neurons to their initial state.& G5\\
$\alpha$ & Number of minimum output spikes for early stopping. & G5\\
$\beta$ & Proportion of non-spiking outputs for early stopping. & G5\\
\end{tabular}
\end{table}

\clearpage

\twocolumn

\bibliographystyle{elsarticle-num} 
\bibliography{biblio_snn}
\end{document}